\documentclass{article}
\usepackage{amsmath}

\usepackage[preprint]{neurips_2024}

\usepackage{pifont}

\usepackage{subcaption}
\usepackage{tcolorbox}
\usepackage{enumitem}
\usepackage{listings}
\tcbuselibrary{breakable, skins, listings}

\usepackage{tcolorbox}
\usepackage{listings}
\tcbuselibrary{breakable, skins, listings}

\usepackage{colortbl}
\definecolor{seablue}{RGB}{219,234,254}

\definecolor{checkgreen}{RGB}{34,139,34}
\definecolor{crossred}{RGB}{200,30,30}
\definecolor{bestgreen}{RGB}{198,239,206}
\definecolor{worstred}{RGB}{255,199,206}

\newcommand{\cmark}{\textcolor{checkgreen}{\ding{51}}}
\newcommand{\xmark}{\textcolor{crossred}{\ding{55}}}

\usepackage[framemethod=tikz]{mdframed}

\newmdenv[
  linecolor=blue!35!black,
  backgroundcolor=blue!2!white,
  linewidth=0.5pt,
  roundcorner=2pt,
  innertopmargin=5pt,
  innerbottommargin=5pt,
  innerleftmargin=6pt,
  innerrightmargin=6pt,
  skipabove=4pt,
  skipbelow=4pt
]{benchmarkframe}

\tcbset{
  benchmarkbox/.style={
    enhanced, breakable,
    colback=gray!5,
    colframe=black!60,
    fonttitle=\bfseries\small,
    coltitle=white,
    attach boxed title to top left={yshift=-2mm, xshift=4mm},
    boxed title style={colback=black!75, sharp corners},
    sharp corners=south,
    top=3mm, bottom=3mm, left=4mm, right=4mm,
    boxrule=0.5pt,
  },
  promptbox/.style={
    enhanced, breakable,
    colback=black!3,
    colframe=black!25,
    boxrule=0.4pt,
    top=2mm, bottom=2mm, left=3mm, right=3mm,
    sharp corners,
    listing only,
    listing options={
      basicstyle=\ttfamily\footnotesize,
      breaklines=true,
      breakatwhitespace=false,
      columns=fullflexible,
      keepspaces=true,
      showstringspaces=false,
    },
  }
}

\usepackage{amsmath}
\usepackage[utf8]{inputenc} 
\usepackage[T1]{fontenc}    
\usepackage{hyperref}       
\usepackage{url}            
\usepackage{graphicx}      
\usepackage{algorithm}   
\usepackage{algpseudocode}
\usepackage{booktabs}       
\usepackage{amsfonts}       
\usepackage{nicefrac}       
\usepackage{microtype}      
\usepackage{xcolor}         
\usepackage{amssymb}
\DeclareMathOperator*{\argmax}{arg\,max}

\usepackage{xspace}
\newcommand{\model}{\textsc{SeaEvo}\xspace}
\newcommand{\modelshinka}{\textsc{SeaEvo}$_\textsc{Shinka}$\xspace}
\newcommand{\modelopen}{\textsc{SeaEvo}$_\textsc{Open}$\xspace}

\usepackage{enumitem}

\usepackage{amssymb}   
\usepackage{booktabs}  
\usepackage{multirow}  

\usepackage{listings}
\usepackage{newunicodechar}

\title{\model: Advancing Algorithm Discovery with Strategy Space Evolution}

\author{%
  Sichun Luo$^1$, Yi Huang$^2$, Haochen Luo$^1$, Fengyuan Liu$^1$, Guanzhi Deng$^3$, Lei Li$^1$\\ \textbf{Qinghua Yao$^1$,  Zefa Hu$^2$, Junlan Feng$^2$, Qi Liu$^1$} \\
  $^1$The University of Hong Kong \quad $^2$JIUTIAN Research, China Mobile \\ $^3$City University of Hong Kong\\
  \texttt{sichunluo2@gmail.com} \\
}

\begin{document}

\maketitle

\begin{abstract}
Large Language Model (LLM)-guided evolutionary search is increasingly used for automated
algorithm discovery, yet most current methods track search progress primarily through
executable programs and scalar fitness. Even when natural-language reasoning is used
through heuristic descriptions or reflection, it typically remains transient mutation context or unstructured memory, rather than organized as persistent
population-level state over strategic directions. As a result, evolutionary search can
struggle to distinguish syntactically different implementations of the same idea, preserve
lower-fitness but strategically promising directions, or detect when an entire family of
strategies has saturated.

We introduce \model, a modular strategy-space layer that turns language-level
strategic reasoning into first-class population-level evolutionary state in LLM-driven
program search. \model represents each candidate program with an explicit natural-language
strategy, clusters the archive by strategy semantics, retrieves behaviorally complementary
inspirations, and periodically navigates the strategy landscape to avoid saturated
directions. 
Without modifying the underlying evolutionary algorithms, \model improves existing evolutionary backbones across algorithm discovery, systems optimization, and agent-scaffold design tasks in most settings.
Across four systems benchmarks, \model achieves a 20.6\% average relative improvement, with the best single run on Prism scoring 3× higher.
These results suggest that persistent strategy representations provide a practical mechanism
for improving the effectiveness and cost-efficiency of LLM-guided evolutionary search,
pointing toward compound AI systems whose search capabilities benefit from the structured
accumulation and reuse of algorithmic strategies.

\end{abstract}

\section{Introduction}

Automatically discovering high-performing algorithms through
Large Language Model (LLM)-driven evolutionary search has emerged as a promising paradigm~\citep{romeraparedes2024funsearch,
novikov2025alphaevolve}. By pairing LLMs as mutation operators with
programmatic evaluators, recent LLM-driven evolutionary systems iteratively propose, execute,
and refine candidate programs, achieving strong results in domains
where search spaces are combinatorially large and hand-crafted
heuristics are costly to design~\citep{liu2024evolution,ye2024reevo,
liu2025cognitive,liu2026systematic}.

\begin{figure}[t]
    \centering
    \includegraphics[width=\linewidth]{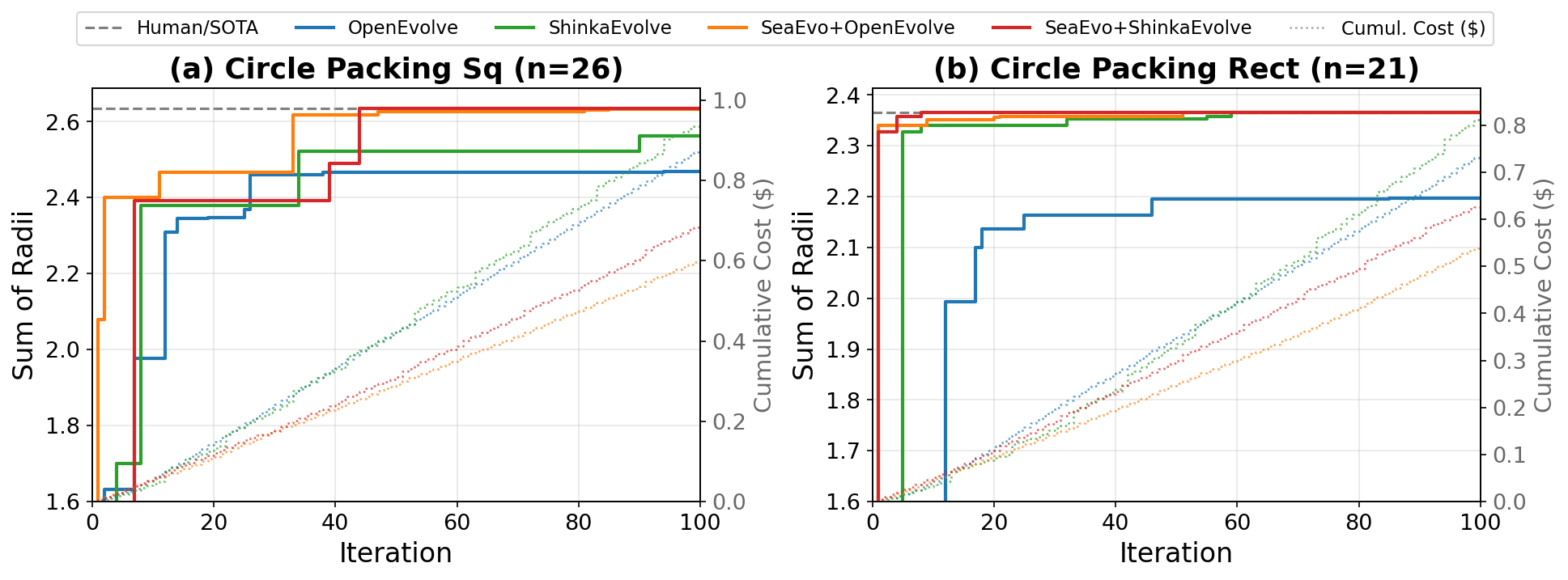}

    \caption{Search trajectories on Circle Packing benchmarks. Solid curves show the 
best-so-far sum of radii over generations, dotted curves show cumulative 
API cost, and the gray dashed line denotes the Human/SOTA reference. 
\model improves both OpenEvolve and ShinkaEvolve backbones, reaching 
competitive solutions earlier while incurring lower cumulative cost. 
}
    \label{fig:search}
    \vspace{-0.15in}
\end{figure}

Despite this progress, most LLM-driven evolutionary systems still 
represent search state primarily as executable programs and scalar 
fitness values \citep{romeraparedes2024funsearch, novikov2025alphaevolve, 
openevolve}. 
Such archives support evaluation and selection, but they provide only a 
limited view of search progress: different programs may instantiate the 
same underlying strategy, similar scores may correspond to qualitatively 
different directions, and low-scoring candidates may encode promising 
ideas that have not yet been refined. 
Recent work has introduced richer language-level signals, including 
natural-language heuristic descriptions \citep{liu2024evolution} and reflection over failures~\citep{shinn2023reflexion, ye2024reevo,agrawal2025gepa}. 
However, these signals are typically used as local prompt context or 
unstructured memory, rather than as a persistent population-level 
representation of semantic strategy families.

This strategy-representation gap creates three practical limitations. 
{First}, syntactically different programs may implement the same underlying idea, 
making redundant variants appear as genuine search progress. 
Second, selection based primarily on scalar fitness may discard lower-scoring candidates 
that encode strategically useful directions or cover complementary failure modes. 
Third, per-program fitness provides little visibility into strategy-family dynamics, 
making it difficult to detect when an entire class of approaches has saturated. 
These limitations suggest that improving LLM-guided evolution requires not only better 
mutation operators or evaluators, but also a richer representation of the evolving 
strategy landscape.


To address this gap, we introduce \textbf{\model} (\underline{\textbf{S}}trat\underline{\textbf{e}}gy-sp\underline{\textbf{a}}ce \underline{\textbf{Evo}}lution), which elevates natural-language strategy to a first-class population-level representation in LLM-driven evolutionary program search.
\model augments each candidate with an explicit natural-language 
strategy description and embedding, yielding a dual-space archive 
organized by both executable programs and semantic strategies. 
It consists of three coordinated modules: {Strategy Articulation 
(SA)}, which diagnoses failures and specifies a strategic direction 
before code generation; {Stratified Experience Retrieval (SER)}, 
which clusters the archive in strategy space and retrieves behaviorally 
complementary inspirations; and {Strategic Landscape Navigation 
(SLN)}, which summarizes effective, saturated, and underexplored 
strategy families to guide future mutations. 
These modules turn the archive from a flat record of programs 
into a navigable map of algorithmic ideas.
We evaluate \model across mathematical algorithm discovery, systems 
optimization, and agent-scaffold design benchmarks, using diverse backbone 
LLMs  and base evolutionary  
frameworks. 
\model improves the underlying backbones in most settings, with the largest gains on open-ended systems optimization tasks where qualitatively distinct strategy families exist. 
Beyond this aggregate gain, \model also improves the search frontier: Figure~\ref{fig:search} shows that it reaches competitive solutions in fewer generations while incurring lower cumulative API cost. 
Case studies in Appendix \ref{sec:case} further show that strategy-space guidance can redirect evolution away from saturated local refinements toward qualitatively different algorithmic families. 
These results suggest that persistent strategy representations provide a practical mechanism for improving the effectiveness and efficiency of LLM-guided evolutionary search.


{\textbf{Contributions.}}
Our contributions are as follows:
1) We identify a {strategy-representation gap} in LLM-guided evolutionary search:
existing archives organize search state around programs and scalar fitness scores, but do
not maintain a population-level semantic representation of the strategies being explored.
2) We introduce \model, a modular strategy-space layer that turns natural-language
strategies into persistent population-level evolutionary state through strategy descriptions,
semantic clustering, behaviorally complementary retrieval, and landscape-level navigation
over effective, saturated, and underexplored strategy families.
3) Extensive experiments across mathematical algorithm discovery,
systems optimization, and agent-scaffold design benchmarks demonstrate the effectiveness of \model. \model improves multiple
open evolutionary backbones in most settings, with a 20.6\% average relative improvement across four
systems
optimization tasks.

\section{Related Work}
\label{sec:related}

\textbf{LLM-Driven Evolutionary Algorithm Discovery.}
Recent work frames algorithm and program improvement as an evolutionary loop in which LLMs act as proposal or mutation operators~\citep{van2024llamea,evox}. FunSearch~\citep{romeraparedes2024funsearch} established this paradigm by pairing an LLM with a programmatic evaluator to evolve mathematical functions, while AlphaEvolve~\citep{novikov2025alphaevolve} scaled it to full codebases using a Gemini ensemble and a MAP-Elites-inspired program database~\citep{mouret2015illuminating}. Subsequent systems improve the efficiency and dynamics of this loop through mechanisms such as adaptive parent sampling and novelty rejection in ShinkaEvolve~\citep{lange2025shinka}, island-based genetic operators in CodeEvolve~\citep{assumpccao2025codeevolve}, hierarchical context management and backtracking in PACEvolve~\citep{yan2026pacevolve}, and adaptive budget and exploration control in AdaEvolve~\citep{cemri2026adaevolve}. These methods primarily improve how programs are sampled, mutated, selected, or orchestrated. \model addresses a complementary axis: what the evolutionary state represents. Rather than replacing the base loop, it augments the archive with persistent strategy descriptions, semantic strategy clusters, and LLM-driven landscape navigation, making it compatible with existing evolutionary backbones.

\textbf{Strategy, Reflection, and Memory in LLM-Based Heuristic Evolution.}
A closely related line of work introduces language-level reasoning into optimization and evolutionary search~\citep{yuksekgonul2025optimizing,lee2025feedback}. EoH~\citep{liu2024evolution} co-evolves natural-language thoughts alongside code; ReEvo~\citep{ye2024reevo}, Reflexion~\citep{shinn2023reflexion}, and GEPA~\citep{agrawal2025gepa} use verbal feedback or reflection to improve future proposals.  Other systems maintain reusable insights~\citep{chen2025hifo}, abstractions \citep{wu2025efficient}, or evolution strategies \citep{yang2025heuragenix}. These works show that natural language can provide useful intermediate structure between scalar feedback and executable code. \model builds on this direction but differs in how the language signal is organized and used: prior systems typically attach thoughts, reflections, insights, or recommendations to individual candidates, lineages, or textual memories, whereas \model organizes the full archive into semantic strategy clusters, retrieves inspirations by behavioral complementarity, and tracks strategy-family dynamics at the population level. This makes strategy space a structured object of search rather than only a source of prompt content.

\begin{figure}[t]
    \centering
    \vspace{-0.18in}
    \includegraphics[width=\linewidth]{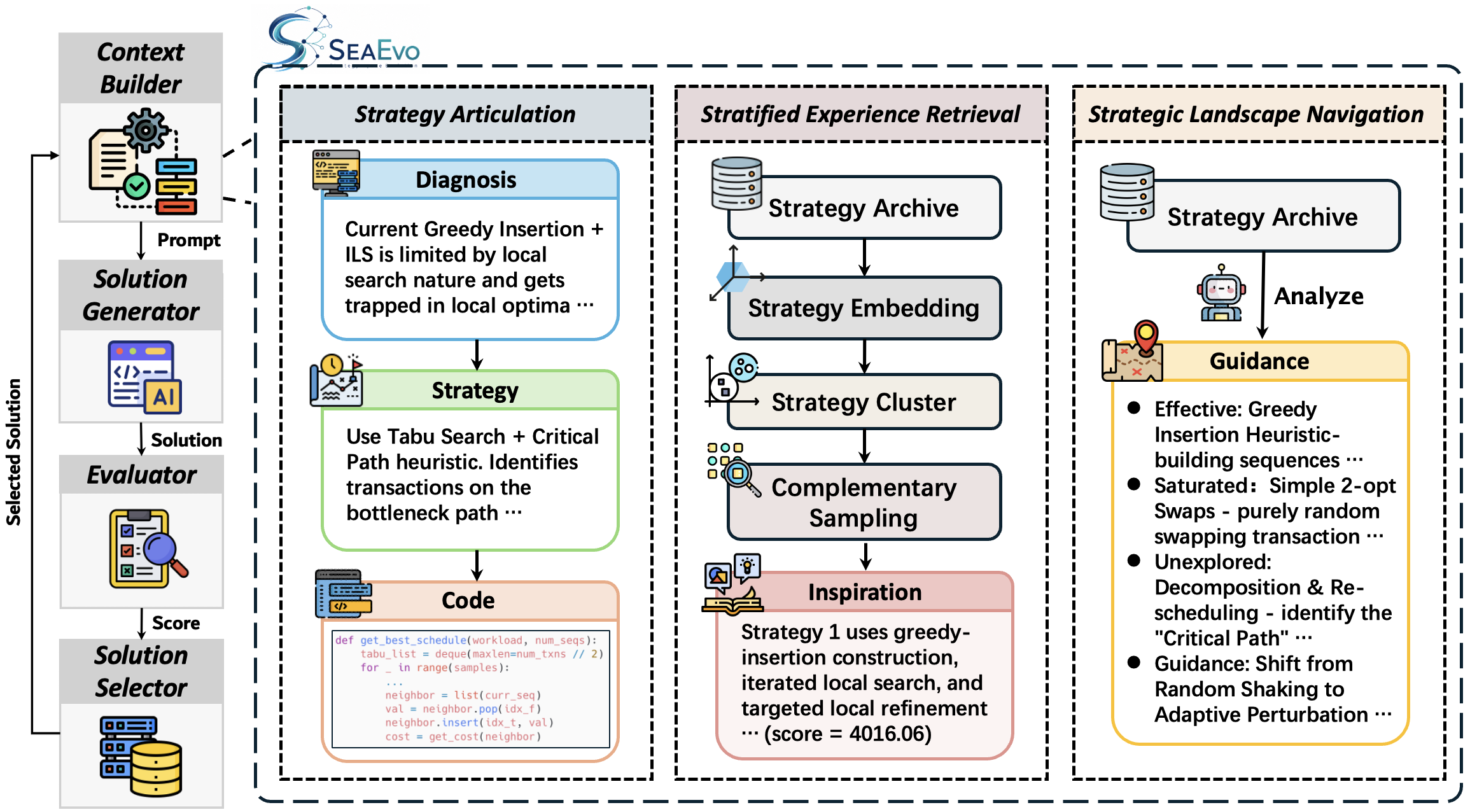}
    \caption{Overview of \model. 
    \model replaces the default context builder with a strategy-aware pipeline, while leaving the base loop's selection and evaluation unchanged.
    Three modules enrich the evolutionary loop: Strategy Articulation replaces blind mutation with a diagnose-direct-implement pipeline; Stratified Experience Retrieval assembles context from semantically clustered, behaviorally diverse archive entries; Strategic Landscape Navigation periodically diagnoses which strategy families are effective, saturated, or underexplored, and guides future mutations accordingly.}
    \label{fig:framework}
    \vspace{-0.18in}
\end{figure}

\section{Method}
\label{sec:method}

\subsection{Problem Formulation}
\label{sec:problem}

We consider the problem of automatically discovering a high-performing 
program for a given task.
Let $f:\mathcal{P}\rightarrow \mathbb{R}$ be an evaluation metric over a 
space of candidate programs $\mathcal{P}$. The goal is to discover
\begin{equation}
    P^* = \argmax_{P \in \mathcal{P}} f(P),
\end{equation}
where $f$ measures task-specific fitness. In an evolutionary framework, 
at generation $t$ an archive $\mathcal{A}_t = \{(P_i, s_i)\}$ with 
$s_i = f(P_i)$ is maintained under a fixed capacity: an LLM prompt is 
constructed from selected entries via context builder, a new candidate is generated and 
evaluated, and then the archive is updated.

Existing frameworks track search progress only through code and scalar 
fitness, or attach natural-language reflections as transient prompt 
context without persisting them at the archive level. Neither representation 
exposes which strategic directions have been explored, which remain 
underrepresented, and which have already saturated. \model addresses 
this by augmenting each archive entry with a persistent strategy 
description and embedding, organizing the archive as a navigable map of 
algorithmic ideas rather than a flat record of programs.

\subsection{Framework Overview}
\label{sec:overview}

Figure~\ref{fig:framework} illustrates \model. The method is a 
modular layer that can be added on top of an existing LLM-driven 
evolutionary framework: it replaces the framework's default context 
builder with a strategy-aware pipeline and augments each archive 
entry with a persistent strategy description and embedding, while 
leaving the base loop's parent selection and evaluation mechanisms unchanged. 
At each generation, the base framework selects a parent program. 
SER retrieves strategy-level inspirations from the strategy-augmented archive; 
SLN provides landscape-level guidance summarizing effective, saturated, and 
underexplored strategy families; and SA uses the parent, retrieved inspirations, 
and landscape guidance to generate a new program together with its strategy 
description. After evaluation, the strategy description is embedded and the new 
entry is inserted into the archive.


The three modules operate at different timescales. SA runs at every 
strategy-guided generation. SER uses a warm-up fallback for the 
first $C$ generations and switches to cluster-based retrieval once 
enough strategy diversity has accumulated. SLN runs every $\Delta$ 
generations and its guidance persists between updates. All modules 
operate through the context construction stage: the base framework 
still sample, evaluates, and selects programs in the usual way, 
while \model changes how mutation context is assembled and what 
strategic state the archive maintains.

\textbf{Exploration ratio.}
To avoid over-constraining the search by the current strategy 
representation, \model uses an $\varepsilon$-greedy scheme: with 
probability $\varepsilon$ it bypasses SA, SER, and SLN and falls back to 
the base framework's default mutation pipeline; with probability 
$1-\varepsilon$ it executes the full \model pipeline. This design balances strategy-guided exploitation with unguided exploration.

\subsection{Strategy Articulation}
\label{sec:sa}

Vanilla LLM mutation often leaves the search direction implicit: the model is
asked to improve a program, but the failure mode and intended algorithmic change
are not explicitly represented~\citep{openevolve}. SA makes this reasoning explicit by asking the
LLM to first articulate a strategic direction before implementing code. Given a
selected parent program $P_{\mathrm{parent}}$, an inspiration set
$\mathcal{G}_t$, and optional landscape guidance $\mathcal{L}_t$, SA produces
a new program $P'$ and a strategy description $d'$ using a single structured
LLM prompt.

The prompt follows a \textit{diagnose--direct--implement} format. 
First, the LLM diagnoses the limitations of $P_{\mathrm{parent}}$ using the 
evaluation signals available to the underlying evolutionary framework, such as 
scalar fitness, execution errors, or task-specific feedback when provided. SA does not require a specific feedback format: in scalar-only
settings, the diagnosis is based on the parent score, the retrieved strategy
inspirations, and the task objective; when richer feedback is available, it can
be included as additional context. Second, conditioned on this diagnosis,
$\mathcal{G}_t$, and optional SLN guidance $\mathcal{L}_t$, the LLM proposes a
concise strategy description $d'$ for the next candidate. When $\mathcal{L}_t$ is
available, it can steer the proposal away from saturated strategy families and
toward underexplored alternatives. Finally, the LLM implements a new program
$P'$ following the proposed strategy description. The diagnosis is used only as
intermediate prompt context. After evaluation, the $P'$ and
$d'$ persist in the archive, with $d'$ embedded for downstream clustering and
retrieval.

This design makes strategy descriptions reusable beyond the immediate mutation
step. Rather than treating reflection as transient prompt content, SA converts
each proposed direction into persistent archive state that can later be clustered
by SER and summarized by SLN.

\subsection{Stratified Experience Retrieval}
\label{sec:ser}

Many evolutionary frameworks construct mutation context by selecting
high-fitness programs or recent ancestors and using their source code as
inspiration for the next mutation~\citep{novikov2025alphaevolve,lange2025shinka}.
While code-level inspiration provides useful implementation details, it is also
costly and noisy: programs can span hundreds of lines, and syntactic similarity
does not necessarily reflect shared algorithmic intent. Moreover, fitness-based
selection can introduce a self-reinforcing bias, in which high-scoring strategy
families are repeatedly surfaced while strategically distinct but temporarily
lower-scoring candidates receive little exposure. SER addresses these issues by
organizing the archive in strategy-embedding space and retrieving inspirations
that cover both similar and distinct strategy
families while remaining behaviorally complementary.

\textbf{Strategy clustering.}
SER partitions the archive into $C$ strategy clusters 
$\{S_1,\ldots,S_C\} \leftarrow k\text{-means}(\{\mathbf{e}_i\}, C)$ 
over strategy embeddings~\citep{lloyd1982least}, recomputed at every generation. 
This groups syntactically different programs that instantiate similar
algorithmic ideas into the same strategy family, providing the semantic
structure for retrieval.
During the warm-up stage ($|\mathcal{A}| < C$), when the archive is too small for reliable 
clustering, SER bypasses the cluster structure and returns the parent, 
the global best, and the program with the highest complementary 
coverage relative to the global best.

\textbf{Complementary Coverage Score.}
SER ranks candidate inspirations by how many of the reference 
program's failures a candidate can solve. 
Given a reference program $P_{\mathrm{ref}}$ and a fixed validation 
set $\mathcal{V}$, each program $P_i$ has a binary success vector 
$\mathbf{b}_i \in \{0,1\}^{|\mathcal{V}|}$. 
We define the \emph{Complementary Coverage score (ComCov)} of $P_i$ with respect to 
$P_{\mathrm{ref}}$ is
\begin{equation}
    \mathrm{ComCov}(P_i;\, P_{\mathrm{ref}})
    = \frac{1}{|\mathcal{V}|}\sum_{k=1}^{|\mathcal{V}|}
      \mathbf{1}\bigl[b_{\mathrm{ref}}^k = 0 \;\wedge\; b_i^k = 1\bigr],
    \label{eq:comcov}
\end{equation}
\textit{i.e.}, the fraction of validation instances that $P_{\mathrm{ref}}$ 
fails but $P_i$ solves. 
ComCov rewards candidates
that cover the reference program's failures, making retrieval complementary to
the parent's weaknesses rather than merely diverse.
When only a scalar fitness $s_i$ is available (\textit{e.g.}, in geometric 
optimization tasks where instance-level success vectors are undefined), 
ComCov is identically zero and ranking falls back to fitness via the 
lexicographic tiebreaker 
$\bigl(\mathrm{ComCov}(P_i;\,\cdot\,),\; s_i\bigr)$.

\textbf{Inspiration selection.}
SER anchors retrieval on the parent program $P_{\mathrm{parent}}$ 
chosen by the base framework. 
After warm-up ($t \geq C$), let $S_{c^*}$ denote the cluster 
containing $P_{\mathrm{parent}}$. 
SER selects two inspirations greedily to maximize marginal 
complementary coverage:
\begin{align}
    P^{\mathrm{intra}}
    &= \argmax_{P_i \in S_{c^*}\setminus\{P_{\mathrm{parent}}\}}
       \mathrm{ComCov}(P_i;\, P_{\mathrm{parent}}),
    \label{eq:intra} \\[4pt]
    c' &\sim \mathrm{Uniform}\{c : c \neq c^*\}, \nonumber \\[2pt]
    P^{\mathrm{cross}}
    &= \argmax_{P_i \in S_{c'}}
       \mathrm{ComCov}_{\Delta}(P_i;\,
         P_{\mathrm{parent}},\, P^{\mathrm{intra}}),
    \label{eq:cross}
\end{align}
where $\mathrm{ComCov}_{\Delta}$ counts only instances failed by 
both $P_{\mathrm{parent}}$ and $P^{\mathrm{intra}}$, ensuring 
non-redundant coverage.
The intra-cluster pick identifies a sibling that shares the parent's 
strategic family but covers different failure cases; the cross-cluster 
pick then targets the residual failures that neither the parent nor 
the sibling can solve.
Random cluster sampling for $c'$, rather than a global argmax, prevents 
the cross-cluster slot from being monopolized by a single dominant 
family, spreading exposure across the strategy landscape over 
successive generations.

For the two retrieved inspirations, SER passes only their strategy 
descriptions and fitness scores, not their source code. 
This reduces the inspiration portion of the prompt by one to two 
orders of magnitude and forces the mutation LLM to reason at the 
level of algorithmic ideas rather than copy code fragments.

\subsection{Strategic Landscape Navigation}
\label{sec:sln}

SER diversifies the local mutation context, but search may still concentrate
within a small number of strategy families. SLN adds a slower-timescale
mechanism that summarizes the population-level strategy landscape and guides
future mutations away from saturated regions.

Every $\Delta$ generations, SLN collects the strategy descriptions and fitness
scores of programs in the current archive and passes them to an LLM. The LLM
produces structured landscape guidance $\mathcal{L}_t$ with four components:
(i)~\emph{effective directions}: strategy families showing consistent
improvement and worth further exploitation;
(ii)~\emph{saturated directions}: families that appear to have plateaued and
should be de-emphasized;
(iii)~\emph{underexplored directions}: plausible approaches not yet well
represented in the archive; and
(iv)~\emph{concrete guidance}: actionable suggestions for subsequent mutation
steps.
The resulting $\mathcal{L}_t$ is appended to the SA prompt and remains active
until the next SLN update.

This design treats saturation and opportunity detection as a semantic
landscape-analysis problem rather than a fixed scalar rule. By reasoning over
strategy descriptions together with their fitness trends, SLN can identify
families that are repeatedly refined without further improvement and suggest
qualitatively different alternatives that would be difficult to recover from
fitness values alone.

\section{Experiments}
\label{sec:experiments}

\subsection{Experiment Setup}

\textbf{Tasks and datasets.}
We evaluate on three task families: (i) four mathematical optimization tasks, including Circle Packing in Square, 
Circle Packing in Rectangle, Heilbronn Triangles, and MinMax 
Distance; (ii) four systems optimization tasks 
from the ADRS benchmark~\citep{adrs}: Prism, TXN, EPLB, 
and LLM-SQL; and (iii) agent-scaffold optimization tasks.

\textbf{Baselines.}
We compare \model against state-of-the-art LLM-driven evolutionary 
methods, including {GEPA}~\citep{agrawal2025gepa}, 
{OpenEvolve}~\citep{openevolve}, and
{ShinkaEvolve}~\citep{lange2025shinka}. 
We instantiate two variants of our method: \modelshinka{} 
over ShinkaEvolve,  and \modelopen{} over OpenEvolve; 
the corresponding backbone rows therefore serve as direct ablations 
of the strategy-aware modules. 
For readability, figures use the expanded form (\textit{e.g.}, \model+ShinkaEvolve) while tables use the compact notation (\textit{e.g.}, \modelshinka{}).
We additionally apply \model 
to GEPA and to AdaEvolve~\citep{cemri2026adaevolve} in 
Appendix~\ref{sec:addtion_exp} to test plug-in generality. 


\textbf{Implementation details.}
We use two backbone LLMs as the mutation model within the evolutionary 
loop: MiMo-V2-Pro~\citep{xiaomi2026mimov2pro} and 
Gemini-3-Flash~\citep{google2025gemini3flash}.
For the agent-scaffold task, 
the inference model is Qwen3-8B~\citep{yang2025qwen3}.
Strategy embeddings use 
{text-embedding-3-small} model~\citep{openai2024textembedding3} . 
We build on the \texttt{SkyDiscover} 
framework~\citep{skydiscover2026}. Each configuration runs for 
$T=100$ generations with SLN update interval $\Delta=10$, $C=5$ 
strategy clusters, and $\varepsilon$-greedy ratio $\varepsilon=0.2$. 
We use the default hyperparameters for all baseline methods.
Similar to \citet{cemri2026adaevolve}, each experiment is repeated three times; we report mean and standard 
deviation.
More details
are provided in Appendix~\ref{sec:appendix_task}.

\begin{table*}[!t]
\caption{Performance on mathematical optimization benchmarks. 
Avg denotes mean score with standard deviation over 3 runs, and Best denotes the best score achieved. 
Best results within each backbone group are bolded, and full-precision near ties are reported in Table~\ref{tab:precision}.
}
\centering
\label{tab:results_math}
\footnotesize
\resizebox{\textwidth}{!}{
\begin{tabular}{lcccccccc}
\toprule
\multirow{2}{*}{\textbf{Method}} & \multicolumn{2}{c}{\textbf{Circle Packing (Rect)}} 
& \multicolumn{2}{c}{\textbf{Circle Packing (Square)}} 
& \multicolumn{2}{c}{\textbf{Heilbronn (Triangles)}} 
& \multicolumn{2}{c}{\textbf{MinMax Distance}} \\
\cmidrule(lr){2-3}
\cmidrule(lr){4-5}
\cmidrule(lr){6-7}
\cmidrule(lr){8-9}
 
& Avg & Best 
& Avg & Best 
& Avg & Best 
& Avg & Best \\



\midrule

\multicolumn{9}{l}{\textit{Backbone: MiMo-V2-Pro}} \\

GEPA         
& 2.3398$_{\scriptsize \pm \text{0.0197}}$   & 2.3584 
& 2.5206$_{\scriptsize \pm \text{0.0207}}$   & 2.5414 
& 0.0265$_{\scriptsize \pm \text{0.0010}}$   & 0.0273 
& 0.2760$_{\scriptsize \pm \text{0.0019}}$ & 0.2776 \\
OpenEvolve   
& 2.1380$_{\scriptsize \pm \text{0.1987}}$   & 2.3530 
& 2.4003$_{\scriptsize \pm \text{0.0386}}$   & 2.4390 
& 0.0250$_{\scriptsize \pm \text{0.0035}}$   & 0.0281 
& 0.2564$_{\scriptsize \pm \text{0.0165}}$ & 0.2760 \\

ShinkaEvolve 
& 2.3560$_{\scriptsize \pm \text{0.0083}}$   & 2.3608 
& 2.4812$_{\scriptsize \pm \text{0.0442}}$   & 2.5123 
& 0.0272$_{\scriptsize \pm \text{0.0001}}$   & 0.0273 
& 0.2776$_{\scriptsize \pm \text{0.0008}}$ & 0.2783 \\

\rowcolor{seablue} \modelopen
& \textbf{2.3607}$_{\scriptsize \pm \text{0.0036}}$   & \textbf{2.3642}
& \textbf{2.6045}$_{\scriptsize \pm \text{0.0338}}$   & \textbf{2.6263}
& \textbf{0.0314}$_{\scriptsize \pm \text{0.0014}}$   & \textbf{0.0330}
& 0.2778$_{\scriptsize \pm \text{0.0003}}$ & 0.2781 \\

\rowcolor{seablue} \modelshinka       
& 2.3603$_{\scriptsize \pm \text{0.0033}}$ & 2.3640
& 2.5467$_{\scriptsize \pm \text{0.1292}}$ & 2.6237
& 0.0274$_{\scriptsize \pm \text{0.0013}}$ & 0.0285
& \textbf{0.2780}$_{\scriptsize \pm \text{0.0003}}$ & \textbf{0.2784} \\

\midrule
\multicolumn{9}{l}{\textit{Backbone: Gemini-3-Flash}} \\

GEPA         
& 2.3610$_{\scriptsize \pm \text{0.0013}}$   & 2.3621 
& 2.5306$_{\scriptsize \pm \text{0.0380}}$   & 2.5574 
& 0.0303$_{\scriptsize \pm \text{0.0011}}$   & 0.0316 
& 0.2785$_{\scriptsize \pm \text{0.0000}}$ & \textbf{0.2785} \\

OpenEvolve   
& 2.1713$_{\scriptsize \pm \text{0.0432}}$   & 2.1968 
& 2.4522$_{\scriptsize \pm \text{0.0387}}$   & 2.4887 
& 0.0062$_{\scriptsize \pm \text{0.0022}}$   & 0.0087 
& 0.2765$_{\scriptsize \pm \text{0.0007}}$ & 0.2774 \\

ShinkaEvolve 
& 2.3625$_{\scriptsize \pm \text{0.0030}}$   & 2.3658 
& 2.5204$_{\scriptsize \pm \text{0.0357}}$   & 2.5617 
& 0.0317$_{\scriptsize \pm \text{0.0012}}$   & 0.0331 
& 0.2783$_{\scriptsize \pm \text{0.0001}}$ & 0.2785 \\

\rowcolor{seablue} \modelopen & \textbf{2.3643}$_{\scriptsize \pm \text{0.0026}}$   & 2.3658
& 2.5802$_{\scriptsize \pm \text{0.0822}}$   & 2.6317
& 0.0305$_{\scriptsize \pm \text{0.0003}}$   & 0.0309
& {0.2785}$_{\scriptsize \pm \text{0.0000}}$ & \textbf{0.2785}\\

\rowcolor{seablue} \modelshinka       
& {2.3639}$_{\scriptsize \pm \text{0.0034}}$ & \textbf{2.3658}
& \textbf{2.6297}$_{\scriptsize \pm \text{0.0077}}$ & \textbf{2.6353}
& \textbf{0.0335}$_{\scriptsize \pm \text{0.0006}}$ & \textbf{0.0342}
& \textbf{0.2785}$_{\scriptsize \pm \text{0.0000}}$ & \textbf{0.2785} \\

\bottomrule
\end{tabular}}
\end{table*}

\begin{table*}[!t]
\caption{Performance on systems optimization benchmarks with Gemini-3-Flash. 
Avg denotes mean score with standard deviation over runs, and Best denotes the best score achieved. 
The Human/SOTA row reports reference best scores when available. 
Best method results are highlighted in boldface.
}
\centering
\label{tab:results_system}
\footnotesize
\resizebox{\textwidth}{!}{
\begin{tabular}{lcccccccc}
\toprule
\multirow{2}{*}{\textbf{Method}} & \multicolumn{2}{c}{\textbf{Prism}} 
& \multicolumn{2}{c}{\textbf{TXN}} 
& \multicolumn{2}{c}{\textbf{EPLB}} 
& \multicolumn{2}{c}{\textbf{LLM-SQL}} \\
\cmidrule(lr){2-3}
\cmidrule(lr){4-5}
\cmidrule(lr){6-7}
\cmidrule(lr){8-9}
 
& Avg & Best 
& Avg & Best 
& Avg & Best 
& Avg & Best \\
\midrule

Human/SOTA&  – &21.89&  – &2725&  –& 0.1265&  – &0.692 

\\

\midrule

GEPA         
& 23.6733$_{\scriptsize \pm \text{0.6034}}$   & 24.0217 
& 3266.0$_{\scriptsize \pm \text{243.1}}$   & 3496.5 
& 0.1472$_{\scriptsize \pm \text{0.0062}}$   & 0.1543 
& 0.7158$_{\scriptsize \pm \text{0.0053}}$ & 0.7233 \\

OpenEvolve   
& 24.0217$_{\scriptsize \pm \text{0.0000}}$   & 24.0217 
& 3586.7$_{\scriptsize \pm \text{188.4}}$   & 3745.3 
& 0.1260$_{\scriptsize \pm \text{0.0002}}$   & 0.1262 
& 0.7117$_{\scriptsize \pm \text{0.0105}}$ & 0.7241 \\

ShinkaEvolve 
& 26.2672$_{\scriptsize \pm \text{0.2757}}$   & 26.5369 
& 3480.7$_{\scriptsize \pm \text{48.6}}$   & 3508.8 
& 0.1326$_{\scriptsize \pm \text{0.0111}}$   & 0.1454 
& 0.6998$_{\scriptsize \pm \text{0.0165}}$ & 0.7181 \\

\rowcolor{seablue} \modelopen 
& 26.1783$_{\scriptsize \pm \text{0.1031}}$   & 26.2783
& 3810.6$_{\scriptsize \pm \text{416.2}}$   & 4098.4
& 0.1535$_{\scriptsize \pm \text{0.0158}}$   & 0.1716
& 0.7104$_{\scriptsize \pm \text{0.0010}}$ & 0.7114 \\

\rowcolor{seablue} \modelshinka       
& \textbf{43.6149}$_{\scriptsize \pm \text{29.7823}}$ & \textbf{78.0040}
& \textbf{3886.1}$_{\scriptsize \pm \text{517.9}}$ & \textbf{4219.4}
& \textbf{0.1958}$_{\scriptsize \pm \text{0.0002}}$ & \textbf{0.1960}
& \textbf{0.7200}$_{\scriptsize \pm \text{0.0087}}$ & \textbf{0.7318} \\

\bottomrule
\end{tabular}}
\end{table*}

\subsection{Main Results}
\label{sec:main-results}

\begin{figure}[t] \vspace{-0.15in}\centering \begin{minipage}{0.48\linewidth} \centering \includegraphics[width=\linewidth]{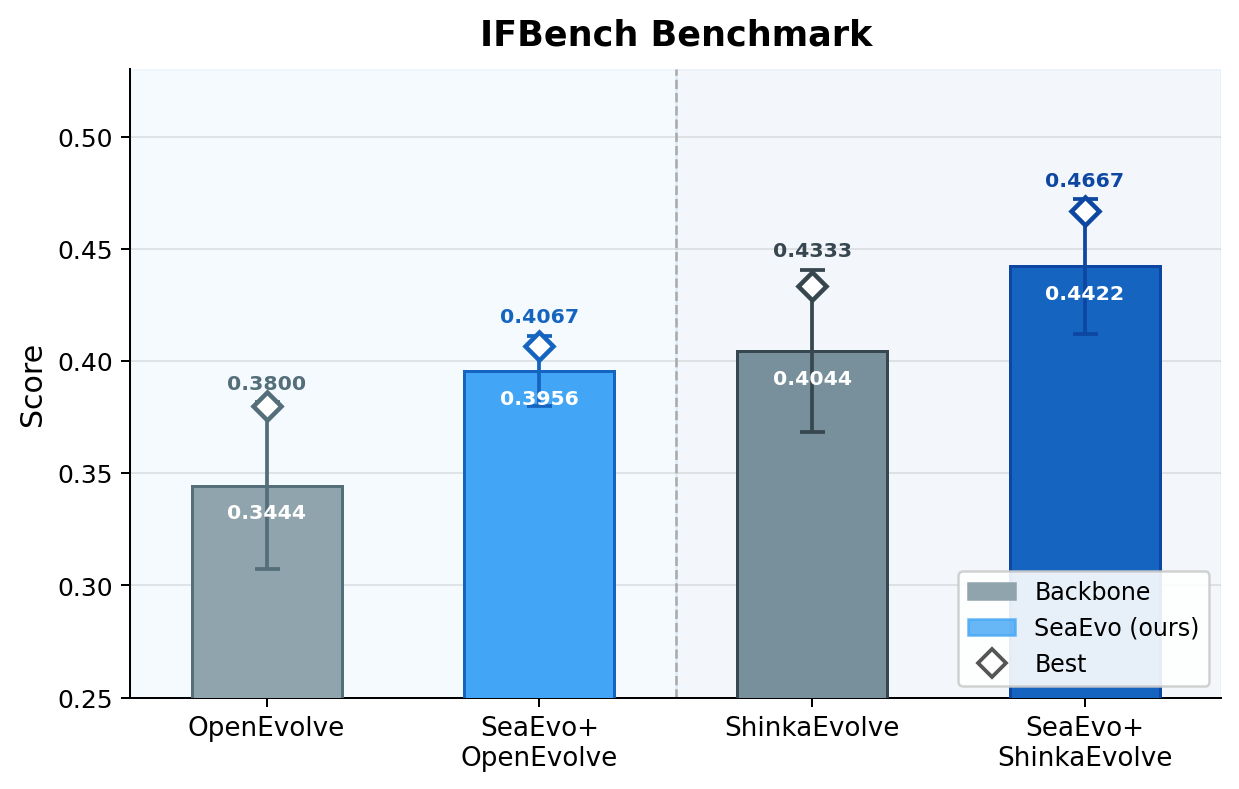} \end{minipage} \hfill \begin{minipage}{0.48\linewidth} \centering \includegraphics[width=\linewidth]{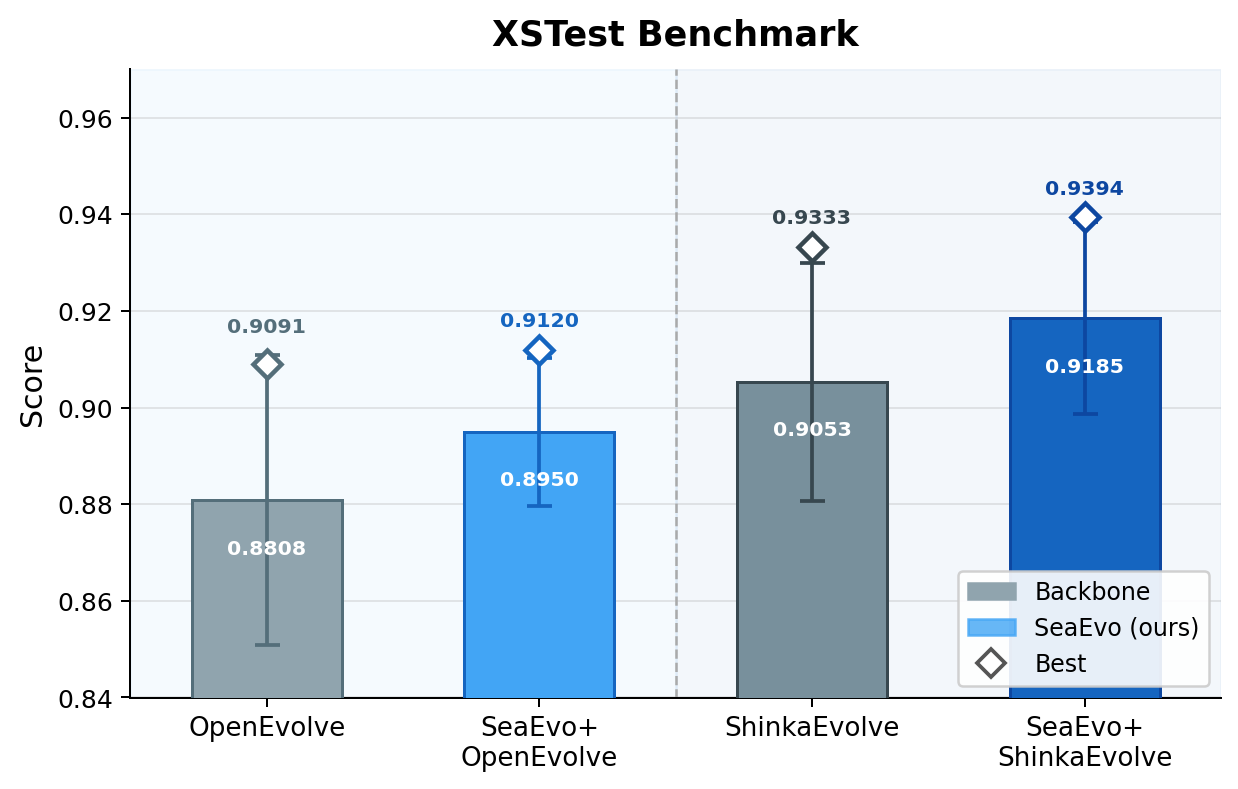} \end{minipage} \caption{Performance on the agent scaffold design benchmark with maximum one LLM call. Bars show average test score with standard deviation across runs, and diamonds indicate the best score. 
} \label{fig:agent} 
\vspace{-0.15in}\end{figure}

Tables~\ref{tab:results_math} and~\ref{tab:results_system} report 
results on mathematical and systems optimization benchmarks, 
respectively. Figure~\ref{fig:agent} and \ref{fig:agent2} report agent-scaffold design results, and Figure~\ref{fig:search} shows search trajectories on the 
two Circle Packing benchmarks. We highlight several observations.

\textbf{Consistent improvement across tasks and backbones.}
Across most configurations in Tables~\ref{tab:results_math} 
and~\ref{tab:results_system}, \model achieves the highest average fitness, and no baseline systematically dominates 
on any task or backbone. The relative ranking is stable between 
MiMo-V2-Pro and Gemini-3-Flash, suggesting 
that gains stem from the strategy-aware modules rather than 
backbone-specific behavior. The same pattern holds on the agent-scaffold design
task (see Figure~\ref{fig:agent} and \ref{fig:agent2}), where \model improves both OpenEvolve 
and ShinkaEvolve backbones. 

\textbf{Gains scale with strategy-space openness.}
\model's advantage is largest on systems optimization benchmarks. \modelshinka{} achieves a 
32.1\% average relative improvement in Avg score over its ShinkaEvolve backbone across the four ADRS benchmarks. On the high-variance Prism benchmark, one run discovers a substantially stronger solution, yielding a near-3× best-score improvement over ShinkaEvolve.
Improvements are more moderate on 
mathematical tasks (\textit{e.g.}, $+4.3\%$ on Circle Packing Square Avg in 
Table~\ref{tab:results_math}), and diminish on narrower-solution-space 
tasks such as LLM-SQL. We attribute this gradient to SLN and SER 
providing stronger signal when qualitatively distinct strategy 
families exist to be discovered and distinguished; when the strategy 
space is constrained, fitness-only baselines are harder to improve 
upon.

\textbf{Faster convergence at lower cost.}
Figure~\ref{fig:search} shows search trajectories on the Circle 
Packing benchmarks. Both \modelopen{} and \modelshinka{} reach the 
Human/SOTA reference earlier than their underlying backbones. On 
Circle Packing (Square), \modelopen{} crosses $2.6$ within $30$ 
generations whereas OpenEvolve plateaus near $2.47$ throughout $100$ 
generations. The cumulative-cost curves show that \model's 
API cost grows slower than the backbones', so the gains do not 
come from additional compute per generation; they come from reaching 
better solutions in fewer generations.

\begin{table}[t]
\centering
\caption{Ablation study on Circle Packing (Square) with Gemini-3-Flash. 
We report mean fitness score (Avg), standard deviation (Std), best score 
(Best), generation at which best was found (Gen@Best), and average 
API cost (AvgCost) across three seeds.
We highlight the best and worst values for 
each metric in green and red, respectively.
}
\label{tab:ablation}
\small
\resizebox{0.7\textwidth}{!}{
\begin{tabular}{ccc|ccccc}
\toprule
\textbf{SA} & \textbf{SER} & \textbf{SLN} 
& \textbf{Avg} & \textbf{Std $\downarrow$} & \textbf{Best} 
& \textbf{Gen@Best $\downarrow$} & \textbf{AvgCost $\downarrow$} \\
\midrule
\xmark & \xmark & \xmark 
& \cellcolor{worstred}2.5023 & 0.0086 
& \cellcolor{worstred}2.5099 & \cellcolor{worstred}99 
& \$0.5367 \\

\cmark & \xmark & \xmark 
& 2.5043 & \cellcolor{worstred}0.0855 & 2.6310 & 67 & \$0.6062 \\

\xmark & \cmark & \xmark 
& 2.5224 & 0.0731 & 2.6310 & 95 & \cellcolor{bestgreen}\$0.5088 \\

\xmark & \xmark & \cmark 
& 2.5634 & 0.0791 & 2.6356 & 61 & \$0.5614 \\

\cmark & \cmark & \xmark 
& 2.5172 & 0.0774 & 2.6304 & 59 & \$0.6576 \\

\cmark & \xmark & \cmark 
& 2.5870 & 0.0816 & 2.6308 & 93 & \cellcolor{worstred}\$0.7147 \\

\xmark & \cmark & \cmark 
& 2.5901 & 0.0724 & \cellcolor{bestgreen}2.6360 & 88 & \$0.5731 \\

\cmark & \cmark & \cmark 
& \cellcolor{bestgreen}{2.6297} 
& \cellcolor{bestgreen}{0.0077} 
& 2.6353 
& \cellcolor{bestgreen}{44} 
& \$0.7075 \\
\bottomrule
\end{tabular}}
\vspace{-0.15in}
\end{table}

\begin{table}[t]
\centering
\caption{Hyperparameter sensitivity on Circle Packing (Square) with 
Gemini-3-Flash. Baseline is the full \model configuration 
($\Delta{=}10$, $k{=}5$, $\varepsilon{=}0.2$). Each row varies one 
hyperparameter while holding the others fixed. The best results are in boldface.}
\label{tab:sensitivity}
\small
\resizebox{0.8\textwidth}{!}{
\begin{tabular}{llccccc}
\toprule
\textbf{Param} & \textbf{Value} & \textbf{Avg} & \textbf{Std $\downarrow$} & \textbf{Best} & \textbf{Gen@Best} $\downarrow$ & \textbf{AvgCost} $\downarrow$ \\
\midrule
\multirow{3}{*}{$\Delta$ (SLN interval)}
 & $\Delta=5$  & 2.5628 & 0.1195 & 2.6343 & 80 & \$0.7710 \\
 & $\Delta=10$ & \textbf{2.6297} & \textbf{0.0077} & \textbf{2.6353} & \textbf{44} & \$0.7075 \\
 & $\Delta=20$ & 2.5742 & 0.0854 & 2.6307 & 74 & \textbf{\$0.6802} \\
\midrule
\multirow{3}{*}{$C$ (clusters)}
 & $C=3$ & 2.6283 & \textbf{0.0052} & 2.6343 & 86 & \textbf{\$0.7006} \\
 & $C=5$ & \textbf{2.6297} & 0.0077 & \textbf{2.6353} & \textbf{44} & \$0.7075 \\
 & $C=8$ & 2.6175 & 0.0173 & 2.6307 & 72 & \$0.7079 \\
\midrule
\multirow{3}{*}{$\varepsilon$ (Exploration ratio)}
 & $\varepsilon=0.0$ & \textbf{2.6341} & \textbf{0.0020} & \textbf{2.6360} & 78 & \$0.7711 \\
 & $\varepsilon=0.2$ & 2.6297 & 0.0077 & 2.6353 & \textbf{44} & \$0.7075 \\
 & $\varepsilon=0.4$ & 2.6263 & 0.0112 & \textbf{2.6360} & 95 & \textbf{\$0.6669} \\
\bottomrule
\end{tabular}}
\vspace{-0.15in}
\end{table}

\subsection{Ablation Study}
\label{sec:ablation}

Table~\ref{tab:ablation} reports ablation results on Circle Packing 
(Square) with Gemini-3-Flash. All three modules improve over the 
no-module baseline, but with distinct roles. SA and SER individually 
yield modest gains in average fitness while substantially raising the 
best fitness, suggesting they primarily expand the reachable solution 
space. SLN alone produces the largest standalone improvement in 
average fitness and the earliest convergence, indicating that 
landscape-level guidance provides the strongest global search signal.
The full \model achieves the best average fitness, finds its best 
solution earliest, and most strikingly reduces variance by an 
order of magnitude relative to every two-module variant, reflecting a 
non-additive synergy among the three modules. Notably, the SA+SER 
configuration underperforms either module alone in average fitness: 
without SLN's landscape signal, structured mutation paired with 
diversity-aware retrieval can over-commit to a narrow set of 
directions. This pattern suggests that SLN's role is not merely 
additive but corrective, it prevents the other two modules from 
locking onto premature strategy families.

\subsection{Hyperparameter Sensitivity}
\label{sec:sensitivity}

Table~\ref{tab:sensitivity} reports sensitivity to three key 
hyperparameters on Circle Packing (Square).

\textbf{SLN update interval $\Delta$.}
$\Delta=10$ achieves the best average fitness and earliest 
convergence. Updating too frequently ($\Delta=5$) destabilizes search: 
the landscape changes before the population has accumulated enough new 
candidates to make the LLM's  judgments 
reliable, producing the highest variance across all settings. 
Updating too infrequently ($\Delta=20$) reduces cost but lets 
ineffective directions persist past the point where they should be 
abandoned, hurting average fitness.

\textbf{Number of clusters $C$.}
Performance is robust around $C=5$, with modest degradation at $C=3$ 
(coarser strategy partitioning conflates distinct families) and $C=8$ 
(too few programs per cluster make intra-cluster retrieval less 
informative). The low sensitivity across this range suggests that 
\model{} does not require careful tuning of $C$.

\textbf{Exploration ratio $\varepsilon$.}
$\varepsilon=0$ (fully strategy-guided, no random exploration) reaches 
the highest best score but does so substantially later than 
$\varepsilon=0.2$, indicating that occasional fallbacks to the base 
mutation pipeline accelerate escape from premature strategy clusters. 
$\varepsilon=0.4$ dilutes the strategy-aware pipeline too aggressively, 
hurting both average fitness and convergence speed. $\varepsilon=0.2$ 
provides the best balance of guided exploration and stochastic 
diversity.

\begin{figure}[t] \centering \begin{minipage}{0.48\linewidth} \centering \includegraphics[width=\linewidth]{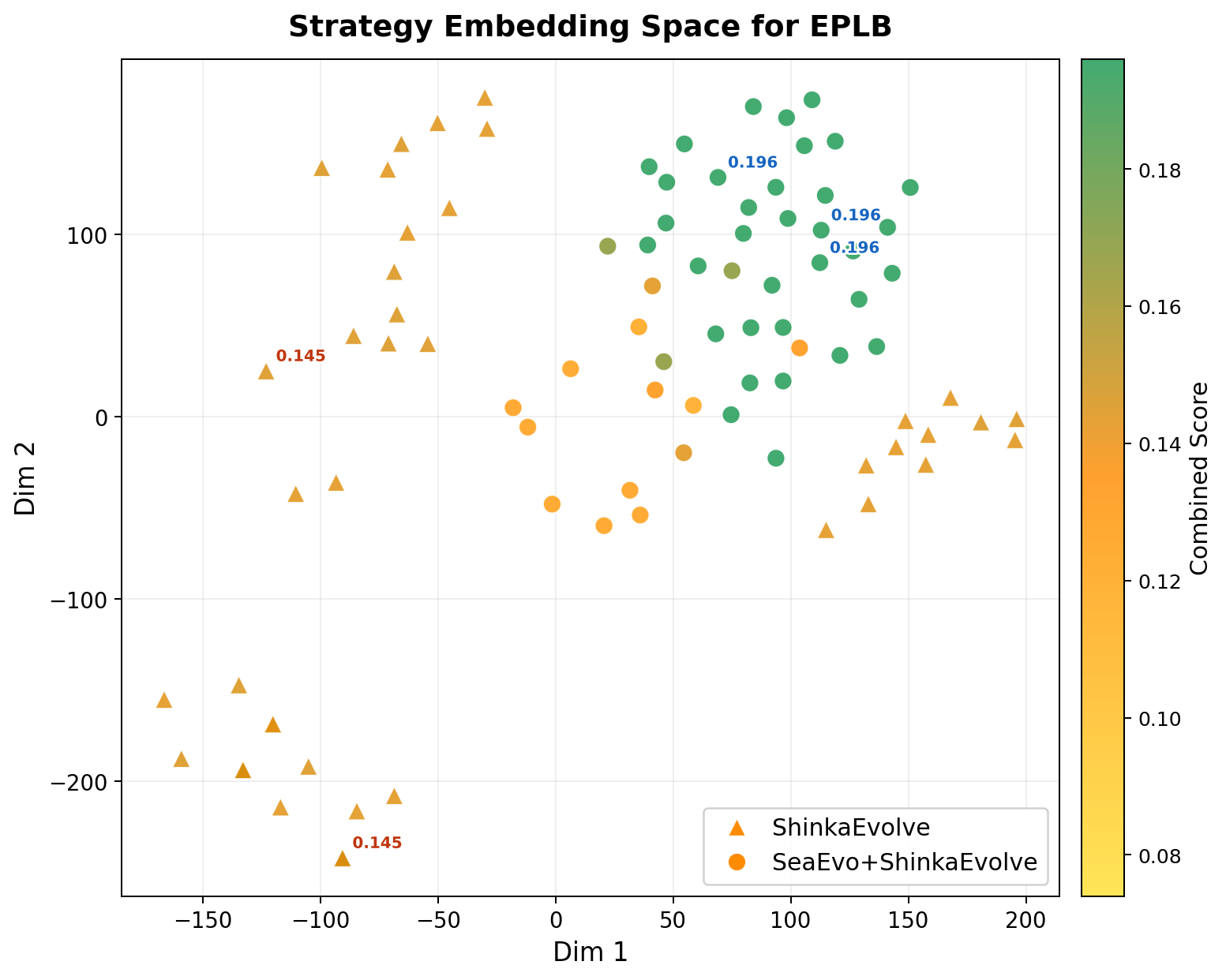} \end{minipage} \hfill \begin{minipage}{0.48\linewidth} \centering \includegraphics[width=\linewidth]{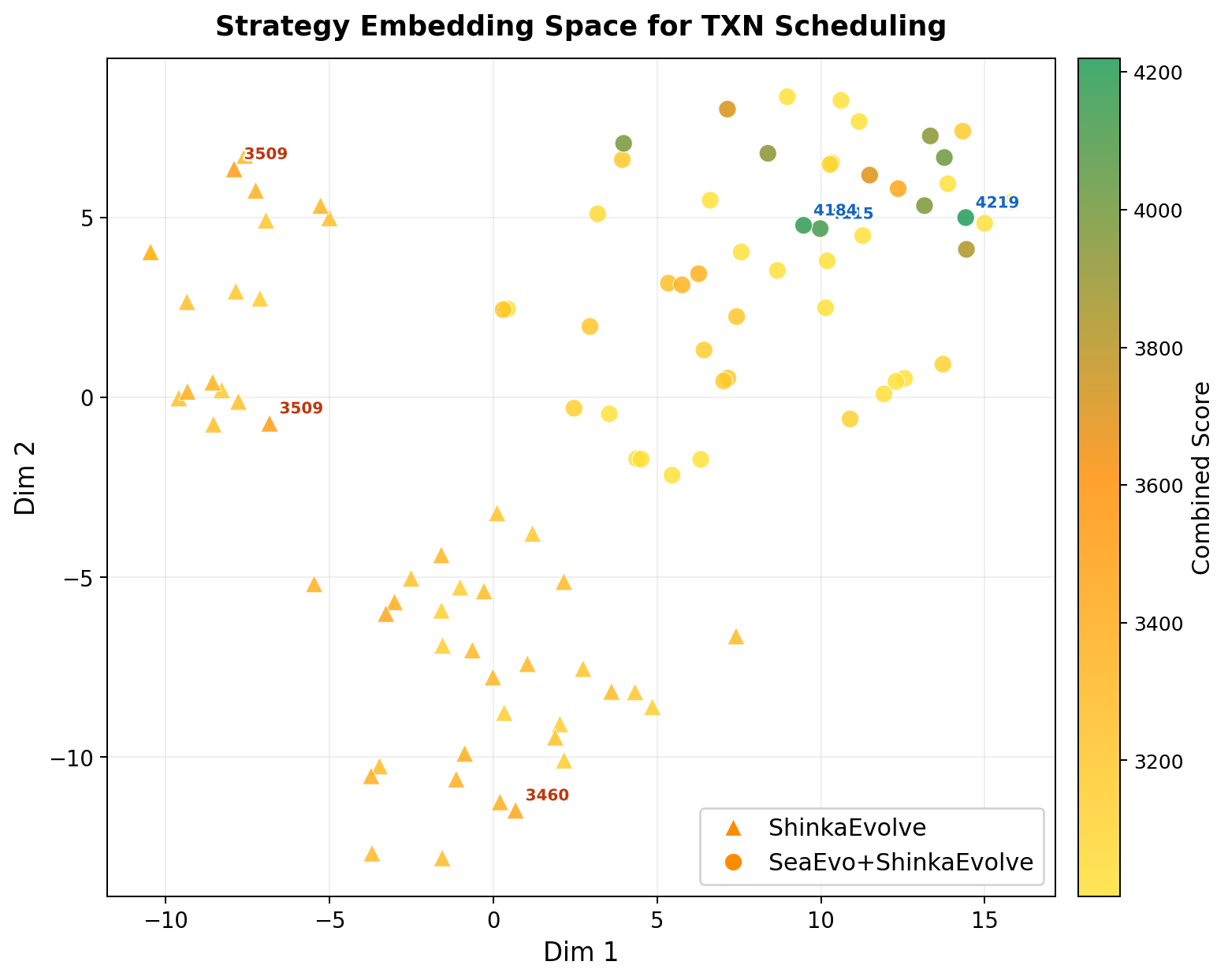} \end{minipage} \caption{Strategy embedding spaces on two systems optimization benchmarks. Each point denotes a candidate program projected from its strategy embedding by t-SNE, color indicates the combined score, and marker shape distinguishes the base backbone from \model-augmented search. } \label{fig:strategy_vis} 
\vspace{-0.15in}
\end{figure}



\subsection{Visualization}
\label{sec:visualization}

To examine whether \model{} changes the distribution of explored strategies rather than merely improving the final score, we project the strategy text associated with all generated programs into two dimensions using t-SNE~\citep{van2008visualizing}. For \modelshinka{}, this text is the explicit strategy description stored in the archive; for the ShinkaEvolve baseline, which does not maintain such descriptions, we approximate it using docstrings and comments extracted from each generated program. Figure~\ref{fig:strategy_vis} shows the resulting qualitative projections on two representative systems optimization benchmarks, EPLB and TXN Scheduling. Each point denotes one generated program; marker shape distinguishes ShinkaEvolve from \modelshinka{}, and color encodes the combined score.
The projections reveal a clear difference in the regions explored by the two search processes. On EPLB, ShinkaEvolve candidates concentrate in a few broad regions, whereas \modelshinka{} produces a dense group of high-scoring candidates in a different part of the strategy space. This suggests that the improvement is associated with discovering and refining a distinct strategy family, rather than only isolated high-scoring mutations. On TXN Scheduling, ShinkaEvolve again concentrates in a small number of regions, while the top \modelshinka{} candidates appear in a separate area of the projection. This is consistent with the intended role of SLN: redirecting mutations toward underexplored strategy families when local directions begin to saturate.

\section{Conclusion}
\label{sec:conclusion}

We introduced \model, a persistent strategy-space layer that turns
language-level strategic reasoning into population-level evolutionary
state. By associating each candidate program with a natural-language
strategy description, clustering the archive by strategy semantics,
retrieving behaviorally complementary inspirations, and monitoring
strategy-family saturation, \model transforms the archive from a flat
record of programs into a navigable map of algorithmic ideas. Across
mathematical algorithm discovery, systems optimization, and agent-scaffold
benchmarks, \model improves existing evolutionary backbones in most
settings, with the largest gains on open-ended tasks. It also improves the
search frontier by reaching competitive solutions in fewer generations
and at lower cumulative API cost. These results point toward LLM-guided
evolutionary systems that improve over time by accumulating, organizing,
and reusing algorithmic strategies.

\bibliographystyle{apalike}
\bibliography{ref}


\appendix

\section{Limitations}
\label{sec:limitations}
We identify several limitations of this work.
First, \model depends on LLMs at multiple stages: generating strategy descriptions in Strategy Articulation, judging which strategy families are saturated or underexplored in Strategic Landscape Navigation, and producing meaningful embeddings for clustering in Stratified Experience Retrieval. If the backbone LLM lacks sufficient domain knowledge, it may mischaracterize promising directions, prematurely abandon useful strategy families, or fail to redirect search away from exhausted ones, which can degrade performance.
Second, due to resource constraints, we evaluate \model with two backbone LLMs across three task families, with each configuration repeated three times. Although the plug-in experiments with GEPA and AdaEvolve provide additional evidence of generality, our experiments do not cover the full range of domains in which LLM-driven evolution may be applied.
Third, \model is most effective when qualitatively distinct strategy families exist and can be meaningfully described in natural language. For tasks with narrow strategy spaces, where the algorithmic design space is highly constrained, natural-language strategy descriptions may fail to capture the relevant axes of variation, limiting the benefit of strategy-space organization.

\section{Broader Impact}
\label{sec:broader-impact}

This work aims to improve LLM-guided evolutionary search by organizing search around persistent natural-language strategy representations. \model may reduce the manual effort required to design optimization heuristics and accelerate scientific, engineering, and systems-optimization workflows where reliable programmatic evaluators are available. It may also make search processes more interpretable by exposing which strategy families have been explored, saturated, or underexplored.

More capable automated program-search systems may also introduce risks if used without appropriate validation. Generated programs can be incorrect, inefficient, brittle, or unsafe outside the benchmark environments in which they are evaluated. In infrastructure, resource-allocation, or safety-critical domains, deploying automatically discovered algorithms without human review could lead to harmful outcomes. 

We mitigate these risks in our experiments through controlled benchmark environments, deterministic evaluators where possible, and sandboxed execution. For practical deployment, we recommend task-specific safety checks, resource limits, robustness testing, and human review before using automatically discovered algorithms in production or safety-critical settings.

\section{Algorithm}

Algorithm~\ref{alg:seaevo} presents the complete \model procedure. 
The outer loop iterates for $T$ generations. 
At each step, SLN refreshes landscape guidance when scheduled 
(line~\ref{alg:sln}); SER assembles a strategy-diverse, behaviorally 
complementary context (line~\ref{alg:ser}); and SA generates a new 
program together with its strategy description (line~\ref{alg:sa}). 
After evaluation, the strategy-augmented archive is updated using the 
base framework's archive update rule (line~\ref{alg:archive}). 
Here, $\mathbf{b}_i$ denotes optional instance-level outcomes when 
available; in scalar-only tasks it can be omitted or treated as empty.

\begin{algorithm}[t]
\caption{\model: Strategy-Space Evolution}
\label{alg:seaevo}
\small
\begin{algorithmic}[1]
\Require Initial program $P_0$, total generations $T$, 
         SLN interval $\Delta$, number of clusters $C$, exploration ratio $\varepsilon$
\Ensure Best program $P^*$
\State Evaluate $P_0$ to obtain $s_0$ and optional $\mathbf{b}_0$;\; 
       $d_0 \leftarrow \textsc{LLM}_{\text{describe}}(P_0)$;\; 
       $\mathbf{e}_0 \leftarrow \text{embed}(d_0)$
\State $\mathcal{A} \leftarrow \{(P_0, s_0, d_0, \mathbf{e}_0, \mathbf{b}_0)\}$;\;
       $P^* \leftarrow P_0$;\; $s^* \leftarrow s_0$;\; 
       $\mathcal{L} \leftarrow \varnothing$
\For{$t = 1, 2, \ldots, T$}
    \State $P_{\text{parent}} \leftarrow \textsc{BaseSelect}(\mathcal{A})$
    \If{$t \bmod \Delta = 0$} \label{alg:sln}
        \State $\mathcal{L} \leftarrow \textsc{SLN}(\mathcal{A})$ 
               \Comment{refresh landscape guidance; persists until next update}
    \EndIf
    \If{$\mathrm{rand}() < \varepsilon$} 
        \Comment{$\varepsilon$-greedy fallback}
        \State $P' \leftarrow \textsc{BaseMutate}(P_{\text{parent}}, \mathcal{A})$;\; 
               $d' \leftarrow \textsc{LLM}_{\text{describe}}(P')$
    \Else
        \State $\mathcal{G}_t \leftarrow \textsc{SER}(\mathcal{A}, P_{\text{parent}}, C)$ 
        \label{alg:ser}
        \State $(P', d') \leftarrow \textsc{SA}(P_{\text{parent}}, \mathcal{G}_t, \mathcal{L})$ 
        \label{alg:sa}
    \EndIf
    \State Evaluate $P'$ to obtain $s'$ and optional $\mathbf{b}'$;\; 
           $\mathbf{e}' \leftarrow \text{embed}(d')$
    \State $\mathcal{A} \leftarrow 
           \textsc{BaseUpdate}(\mathcal{A}, (P', s', d', \mathbf{e}', \mathbf{b}'))$ 
           \label{alg:archive}
    \If{$s' > s^*$}
        \State $P^* \leftarrow P'$;\; $s^* \leftarrow s'$
    \EndIf
\EndFor
\State \Return $P^*$
\end{algorithmic}
\end{algorithm}

\begin{algorithm}[t]
\caption{\textsc{SA}: Strategy Articulation}
\label{alg:sa-detail}
\small
\begin{algorithmic}[1]
\Require Parent program $P_{\text{parent}}$, inspiration set $\mathcal{G}_t$, landscape guidance $\mathcal{L}$
\Ensure New program $P'$, strategy description $d'$
\State $(\text{diag},\, d',\, P') 
       \leftarrow \textsc{LLM}_{\text{SA}}(P_{\text{parent}},\, \mathcal{G}_t,\, \mathcal{L})$
       \Comment{single call: diagnose $\rightarrow$ direct $\rightarrow$ implement}
\State \Return $P',\, d'$
\end{algorithmic}
\end{algorithm}

\begin{algorithm}[t]
\caption{\textsc{SER}: Stratified Experience Retrieval}
\label{alg:ser-detail}
\small
\begin{algorithmic}[1]
\Require Archive $\mathcal{A}$, parent $P_{\text{parent}}$, number of clusters $C$
\Ensure Inspiration context $\mathcal{G}_t$
\State $P_{\text{best}} \leftarrow \argmax_{P_i \in \mathcal{A}} s_i$
\If{$|\mathcal{A}| < C$} 
    \Comment{warm-up fallback before reliable clustering}
    \State $\mathcal{B} \leftarrow 
           \mathcal{A} \setminus \{P_{\text{parent}}, P_{\text{best}}\}$
    \If{$\mathcal{B} \neq \varnothing$}
        \State $P^{\text{div}} \leftarrow 
               \argmax_{P_i \in \mathcal{B}} \text{score}(P_i;\, P_{\text{best}})$
        \State \Return unique entries from 
               $\{P_{\text{parent}},\, P_{\text{best}},\, P^{\text{div}}\}$
    \Else
        \State \Return unique entries from 
               $\{P_{\text{parent}},\, P_{\text{best}}\}$
    \EndIf
\EndIf
\State $\{S_c\}_{c=1}^C \leftarrow 
       \textsc{KMeans}(\{\mathbf{e}_i : 
       (P_i, s_i, d_i, \mathbf{e}_i, \mathbf{b}_i) \in \mathcal{A}\}, C)$
\State $c^* \leftarrow$ cluster containing $P_{\text{parent}}$
\If{$S_{c^*} \setminus \{P_{\text{parent}}\} = \varnothing$}
    \State $P^{\text{intra}} \leftarrow 
           \argmax_{P_i \in \mathcal{A} \setminus \{P_{\text{parent}}\}} 
           \text{score}(P_i;\, P_{\text{parent}})$
\Else
    \State $P^{\text{intra}} \leftarrow 
           \argmax_{P_i \in S_{c^*} \setminus \{P_{\text{parent}}\}} 
           \text{score}(P_i;\, P_{\text{parent}})$
\EndIf
\State $c' \sim \text{Uniform}\{c : c \neq c^*\}$ 
       \Comment{random cross-cluster sampling}
\State $P^{\text{cross}} \leftarrow 
       \argmax_{P_i \in S_{c'}} 
       \text{score}_{\Delta}(P_i;\, P_{\text{parent}}, P^{\text{intra}})$
\State \Return $\{P_{\text{parent}},\, P^{\text{intra}},\, P^{\text{cross}}\}$
\end{algorithmic}
\end{algorithm}

\begin{algorithm}[t]
\caption{\textsc{SLN}: Strategic Landscape Navigation}
\label{alg:sln-detail}
\small
\begin{algorithmic}[1]
\Require Archive $\mathcal{A}$
\Ensure Landscape guidance $\mathcal{L}$
\State $\mathcal{D} \leftarrow 
       \{(d_i, s_i) : (P_i, s_i, d_i, \mathbf{e}_i, \mathbf{b}_i) \in \mathcal{A}\}$
       \Comment{collect strategy descriptions and fitness scores}
\State $\mathcal{L} \leftarrow \textsc{LLM}_{\text{navigate}}(\mathcal{D})$
       \Comment{identify effective, saturated, and underexplored families}
\State \Return $\mathcal{L}$
\end{algorithmic}
\end{algorithm}

\section{Experiment Setup}
\label{sec:appendix_task}

\subsection{Datasets and Tasks}

For mathematical optimization, we evaluate on four algorithm discovery tasks:
Circle Packing (Rectangle), Circle Packing (Square), Heilbronn Triangles, and
MinMax Distance. 
For systems optimization, we use four tasks from the ADRS benchmark~\citep{adrs}:
Prism, TXN, EPLB, and LLM-SQL. 
The detailed task descriptions and prompts are provided in
Appendix~\ref{sec:prompt}.

For agent-scaffold design, we use the IFBench \citep{ifbench} and XSTest benchmark~\citep{rottger2024xstest}. 
We evaluate each benchmark under two inference-budget settings, where the inference model is allowed to make either one or up to three LLM calls per example.
We split the benchmark into a feedback set, a metric set, and a test set, with
60, 75, and 150 examples, respectively. 
The feedback set is used during evolution, the metric set is used for candidate
selection, and the test set is held out for final evaluation.

\subsection{Models}

For evolution, we adopt two frontier models as representative non-reasoning and
reasoning backbones: Gemini-3-Flash~\citep{google2025gemini3flash}, accessed via the
OpenRouter API\footnote{\url{https://openrouter.ai/api/v1}}, and MiMo-V2-Pro~\citep{xiaomi2026mimov2pro}, accessed via its
official API\footnote{\url{https://api.xiaomimimo.com/v1}}. 
The decoding temperature for evolution model is set to 0.7.
For agent-scaffold tasks, we use Qwen3-8B~\citep{yang2025qwen3} as the
inference model, deployed locally with vLLM~\citep{kwon2023efficient} on
2$\times$A6000 GPUs. 
Following \citet{li2025c}, the default maximum generation length is set to 4096 and the decoding temperature for the inference model is set to 1.0. The decoding temperature can be modified by the evolution model.





We compare \model{} against several representative LLM-driven evolutionary methods and use the corresponding open-source backbones to instantiate \model{}.
{GEPA}~\citep{agrawal2025gepa}  
performs reflective 
prompt evolution using verbal feedback over execution traces. 
{OpenEvolve}~\citep{openevolve} is an open-source 
reimplementation of the AlphaEvolve~\citep{novikov2025alphaevolve} 
paradigm. {ShinkaEvolve}~\citep{lange2025shinka} improves 
sample efficiency via adaptive parent sampling and novelty-based 
rejection. 
AlphaEvolve \citep{novikov2025alphaevolve} is closed-source and is 
included only as a reference target. 
We use default hyperparameters as reported in the respective papers.
For {OpenEvolve}, we set \texttt{num\_context\_programs}$=5$,
\texttt{num\_islands}$=5$, \texttt{population\_size}$=40$,
\texttt{archive\_size}$=100$, \texttt{exploration\_ratio}$=0.2$,
\texttt{exploitation\_ratio}$=0.7$, \texttt{migration\_interval}$=10$,
\texttt{migration\_rate}$=0.1$, and \texttt{feature\_bins}$=10$.
For {ShinkaEvolve}, we set \texttt{num\_islands}$=3$,
\texttt{archive\_size}$=20$, \texttt{migration\_interval}$=5$,
and \texttt{migration\_rate}$=0.2$.

\section{Additional Experimental Results}
\label{sec:addtion_exp}

\begin{table*}[t]
\caption{Generality of \model{} as a plug-in over additional 
evolutionary backbones GEPA and AdaEvolve, evaluated on a mix of 
mathematical and systems 
optimization benchmarks. Within each backbone group, 
\model{}-augmented rows (highlighted) are compared against the 
corresponding base framework. Best results  are in boldface.
}
\centering
\label{tab:results}
\footnotesize
\resizebox{\textwidth}{!}{
\begin{tabular}{lcccccccc}
\toprule
\multirow{2}{*}{\textbf{Method}} 
& \multicolumn{2}{c}{\textbf{Circle Packing (Rect)}} 
& \multicolumn{2}{c}{\textbf{Heilbronn (Triangles)}} 
& \multicolumn{2}{c}{\textbf{TXN}} 
& \multicolumn{2}{c}{\textbf{EPLB}} \\
\cmidrule(lr){2-3}
\cmidrule(lr){4-5}
\cmidrule(lr){6-7}
\cmidrule(lr){8-9}
 
& Avg & Best 
& Avg & Best 
& Avg & Best 
& Avg & Best \\
\midrule

GEPA         
& 2.3610$_{\scriptsize \pm 0.0013}$ & 2.3621
& 0.0303$_{\scriptsize \pm 0.0011}$ & 0.0316
& 3266.0$_{\scriptsize \pm 243.1}$  & 3496.5 
& 0.1472$_{\scriptsize \pm 0.0062}$ & 0.1543 \\

\rowcolor{seablue} 
\quad + \model 
& 2.5408$_{\scriptsize \pm 0.0442}$ & 2.6022
& 0.0322$_{\scriptsize \pm 0.0017}$ & 0.0337
& 3386.0$_{\scriptsize \pm 11.0}$   & 3401.0
& 0.1397$_{\scriptsize \pm 0.0178}$ & 0.1649 \\

AdaEvolve   
& 2.6223$_{\scriptsize \pm 0.0086}$ & 2.6285 
& 0.0322$_{\scriptsize \pm 0.0003}$ & 0.0324
& 3622.0$_{\scriptsize \pm 305.0}$  & 4032.0
& 0.1270$_{\scriptsize \pm 0.0002}$ & 0.1273 \\

\rowcolor{seablue} 
\quad + \model 
& \textbf{2.6283}$_{\scriptsize \pm 0.0097}$ & \textbf{2.6360}
& \textbf{0.0329}$_{\scriptsize \pm 0.0014}$ & \textbf{0.0338}
& \textbf{4238.0}$_{\scriptsize \pm 598.0}$  & \textbf{5076.0}
& \textbf{0.1597}$_{\scriptsize \pm 0.0111}$ & \textbf{0.1689} \\

\bottomrule
\end{tabular}}
\end{table*}

\subsection{Generality across additional evolutionary frameworks.}
To examine whether \model{} is tied to a specific evolutionary 
backbone, we additionally integrate it into two further LLM-driven 
frameworks: GEPA~\citep{agrawal2025gepa} and 
AdaEvolve~\citep{cemri2026adaevolve}. Table~\ref{tab:results} shows 
that \model{} can be plugged into both backbones with no architectural 
changes and improves their average score on most benchmarks; the gains 
on AdaEvolve are particularly large on the open-ended systems tasks 
(TXN and EPLB), consistent with the strategy-space-openness pattern 
observed in the main results. The one exception is EPLB under GEPA, 
where \model{} slightly underperforms the backbone in Avg but improves 
Best---suggesting that GEPA's reflective prompt evolution may already 
saturate parts of the strategy signal \model{} would otherwise add. 
Overall, these results indicate that \model{} acts as a plug-and-play 
strategy-space layer that generalizes beyond the OpenEvolve and 
ShinkaEvolve backbones used in our main experiments.

\begin{table}[t]
\centering
\caption{Ablation study on TXN with Gemini-3-Flash. 
We report mean fitness (Avg), standard deviation (Std), best score 
(Best), generation at which best was found (Gen@Best), and average 
API cost (AvgCost) across three runs. We highlight the best and worst values for each
metric in green and red, respectively.}
\label{tab:ablation2}
\small
\begin{tabular}{ccc|ccccc}
\toprule
\textbf{SA} & \textbf{SER} & \textbf{SLN} 
& \textbf{Avg} & \textbf{Std $\downarrow$} & \textbf{Best} 
& \textbf{Gen@Best $\downarrow$} & \textbf{AvgCost $\downarrow$} \\
\midrule
\xmark & \xmark & \xmark 
& \cellcolor{worstred}3541.4 & 163.5 & \cellcolor{worstred}3676.5 & 53 & 0.5055 \\

\cmark & \xmark & \xmark 
& 3668.1 & \cellcolor{bestgreen}136.9 & 3831.4 & 94 & 0.6115 \\

\xmark & \cmark & \xmark 
& 3599.8 & 206.0 & 3891.1 & \cellcolor{worstred}97 & \cellcolor{bestgreen}0.4811 \\

\xmark & \xmark & \cmark 
& 3881.1 & 142.9 & 4081.6 & 94 & 0.5605 \\

\cmark & \cmark & \xmark 
& 3653.4 & 170.4 & 3787.9 & 85 & 0.6205 \\

\cmark & \xmark & \cmark 
& 3784.9 & 317.0 & \cellcolor{bestgreen}4219.4 & 89 & 0.6884 \\

\xmark & \cmark & \cmark 
& 3737.7 & 351.4 & 4132.2 & \cellcolor{bestgreen}30 & 0.5711 \\

\cmark & \cmark & \cmark 
& \cellcolor{bestgreen}3886.1 & \cellcolor{worstred}517.9 & \cellcolor{bestgreen}4219.4 & 59 & \cellcolor{worstred}0.6966 \\
\bottomrule
\end{tabular}
\end{table}

\begin{table}[t]
\centering
\caption{Full-precision values for results that appear tied after rounding in Table~\ref{tab:results_math}, including Circle Packing (Rectangle, $n{=}21$) and MinMax Distance.
}
\label{tab:precision}
\small
\begin{tabular}{llllc}
\toprule
\textbf{Task} & \textbf{Metric} & \textbf{Backbone} & \textbf{Method} & \textbf{Exact Number} \\
\midrule

\multirow{3}{*}{Circle Packing (Rect)} 
& \multirow{3}{*}{Best}
& \multirow{3}{*}{Gemini-3-Flash}
    & ShinkaEvolve        & 2.365832354910838 \\
&     &     
    & \modelopen   & 2.365832376111432 \\
&     &     
    & \modelshinka & \textbf{2.365832393877214} \\

\midrule

\multirow{8}{*}{MinMax Distance}
& \multirow{5}{*}{Best}
& MiMo-V2-Pro
& \modelopen
& 0.278493639030704 \\
\cmidrule(lr){3-5}

& & \multirow{4}{*}{Gemini-3-Flash}
& GEPA
& \textbf{0.278539322443734} \\
& &
& ShinkaEvolve
& 0.278539278338067 \\
& &
& \modelopen
& \textbf{0.278539322443734} \\
& &
& \modelshinka
& \textbf{0.278539322443734} \\

\cmidrule(lr){2-5}

& \multirow{3}{*}{Mean}
& \multirow{3}{*}{Gemini-3-Flash}
    & GEPA                & 0.278536175170671 \\
&     &
    & \modelopen          & 0.278539322443661 \\
&     &
    & \modelshinka        & \textbf{0.278539322443734} \\

\bottomrule
\end{tabular}
\end{table}


\begin{figure}[t]
    \centering
    \begin{minipage}{0.48\linewidth}
        \centering
        \includegraphics[width=\linewidth]{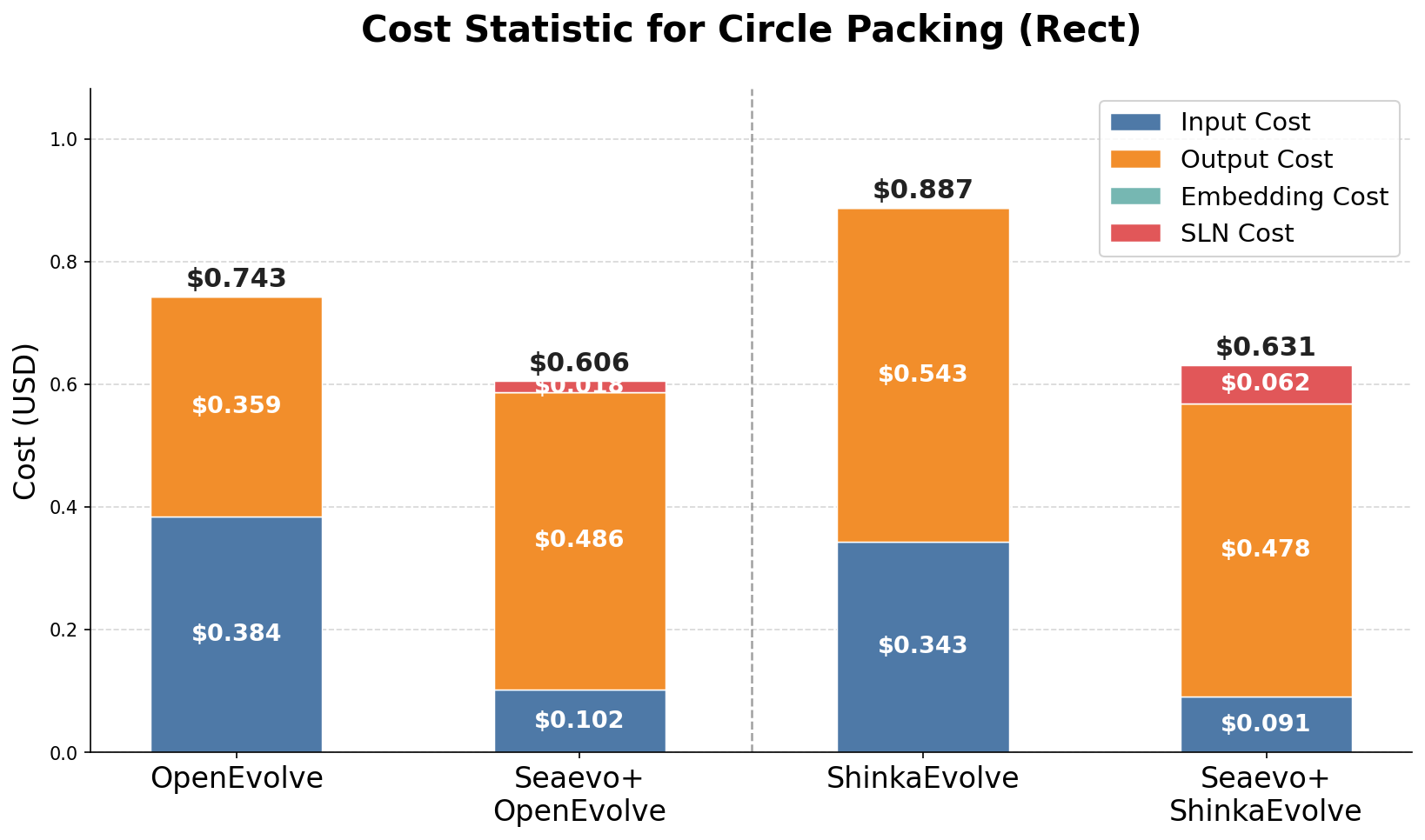}
    \end{minipage}
    \hfill
    \begin{minipage}{0.48\linewidth}
        \centering
        \includegraphics[width=\linewidth]{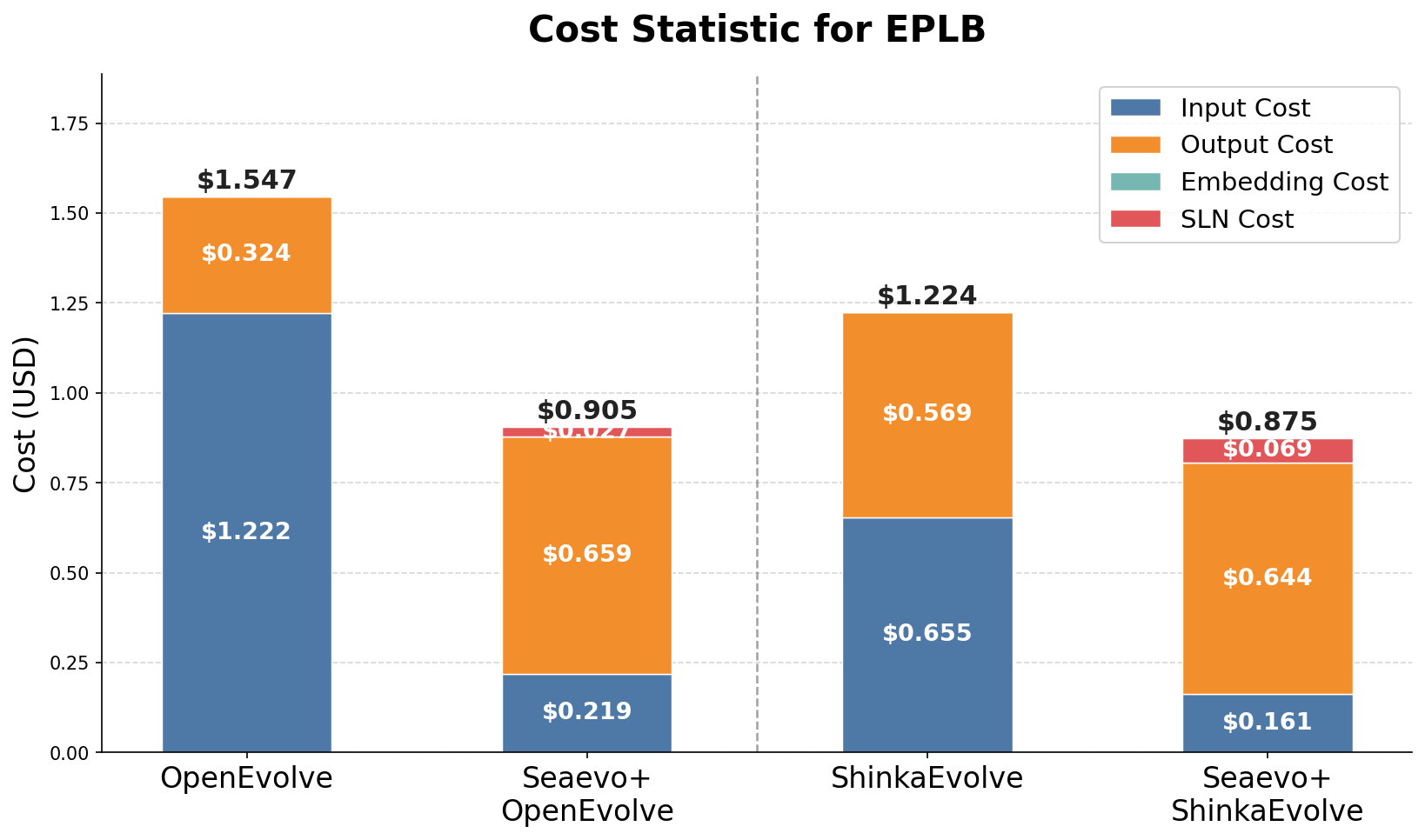}
    \end{minipage}
    \caption{Cost breakdown on Circle Packing (Rect) and EPLB
with Gemini-3-Flash.  Bars are decomposed into input cost (prompt
tokens), output cost (completion tokens), embedding cost
(\texttt{text-embedding-3-small} calls for strategy descriptions),
and SLN cost (landscape navigation for strategy navigation).
\model reduces total cost in all four comparisons despite
adding embedding and SLN overhead, primarily by replacing source-code
context with compact strategy descriptions in the mutation prompt.}
    \label{fig:cost}
\end{figure}

\subsection{Ablation Study on TXN}

Table~\ref{tab:ablation2} shows that the full \model framework achieves the
strongest overall performance, obtaining the best average fitness and matching
the highest best score among all variants. This indicates that the three modules
are complementary: strategy articulation provides more deliberate mutations,
stratified retrieval improves the diversity of useful inspirations, and landscape
navigation helps steer the search away from over-explored directions.

Among the individual modules, SLN is the most effective standalone component,
nearly matching the full model in average fitness. This suggests that
strategy-level landscape guidance is particularly important for TXN, where the
search can otherwise spend many generations refining already saturated
directions. In contrast, SA and SER alone bring more limited gains. SA improves
the stability of mutation by making the intended improvement explicit, while SER
mainly broadens the inspiration context. However, neither is sufficient on its
own to consistently guide the global search.

The interaction between modules reveals a trade-off between early discovery and 
overall performance. SER+SLN finds a strong solution the earliest, indicating 
that complementary retrieval can accelerate landscape-guided exploration. 
However, the full model achieves the best overall performance, suggesting that 
SA provides additional mutation-level guidance that improves the quality of the 
search. This comes with higher variance and cost, reflecting a more aggressive 
and exploratory search process.
Overall, these results show that \model's advantage does not come from a single
component in isolation, but from coupling local mutation-level reasoning with
archive-level retrieval and landscape-level search control.

\subsection{Cost Analysis}
\label{app:cost}
 
Figure~\ref{fig:cost} reports the per-run API cost breakdown on
Circle Packing (Rect) and EPLB. Across both tasks and both evolutionary
backbones, \model consistently reduces total cost, with savings
ranging from 18.4\% to 41.5\%. The reduction is most pronounced on EPLB,
where candidate programs are longer and code-level context becomes
substantially more expensive.

The main source of savings is lower input-token cost. Instead of passing
full source code for retrieved inspiration programs, SER provides compact
strategy descriptions together with fitness information. This replacement
substantially shortens the mutation prompt, especially for tasks with
longer programs. Output cost remains comparable, and can slightly increase
because SA asks the model to produce structured diagnose--direct--implement
reasoning before generating the new candidate.

The additional overhead introduced by strategy embeddings and SLN is small.
Embedding calls are inexpensive, and SLN is invoked only periodically, so
these components do not offset the savings from shorter prompts. Overall,
the results suggest that strategy-level context is not only more informative
for search, but also more token-efficient than code-level context. This
advantage becomes more important as the evolved programs grow longer.

\subsection{Additional Prism Runs}

Prism exhibits substantially higher variance than the other ADRS benchmarks. To better characterize this behavior, we ran three additional independent \modelshinka trials on Prism. The results are shown in Table~\ref{tab:additional-prism-runs}. Across these additional runs, \modelshinka obtains an average score of $32.99$ with a sample standard deviation of $8.79$.

\begin{table}[t]
\centering
\caption{Additional Prism results over three independent runs.}
\label{tab:additional-prism-runs}
\begin{tabular}{lcccc}
\toprule
 & R1 & R2 & R3 & Mean $\pm$ Std \\
\midrule
Prism Score & 26.26 & 42.94 & 29.77 & $32.99 \pm 8.79$ \\
\bottomrule
\end{tabular}
\end{table}

\begin{figure}[t] \centering \begin{minipage}{0.48\linewidth} \centering \includegraphics[width=\linewidth]{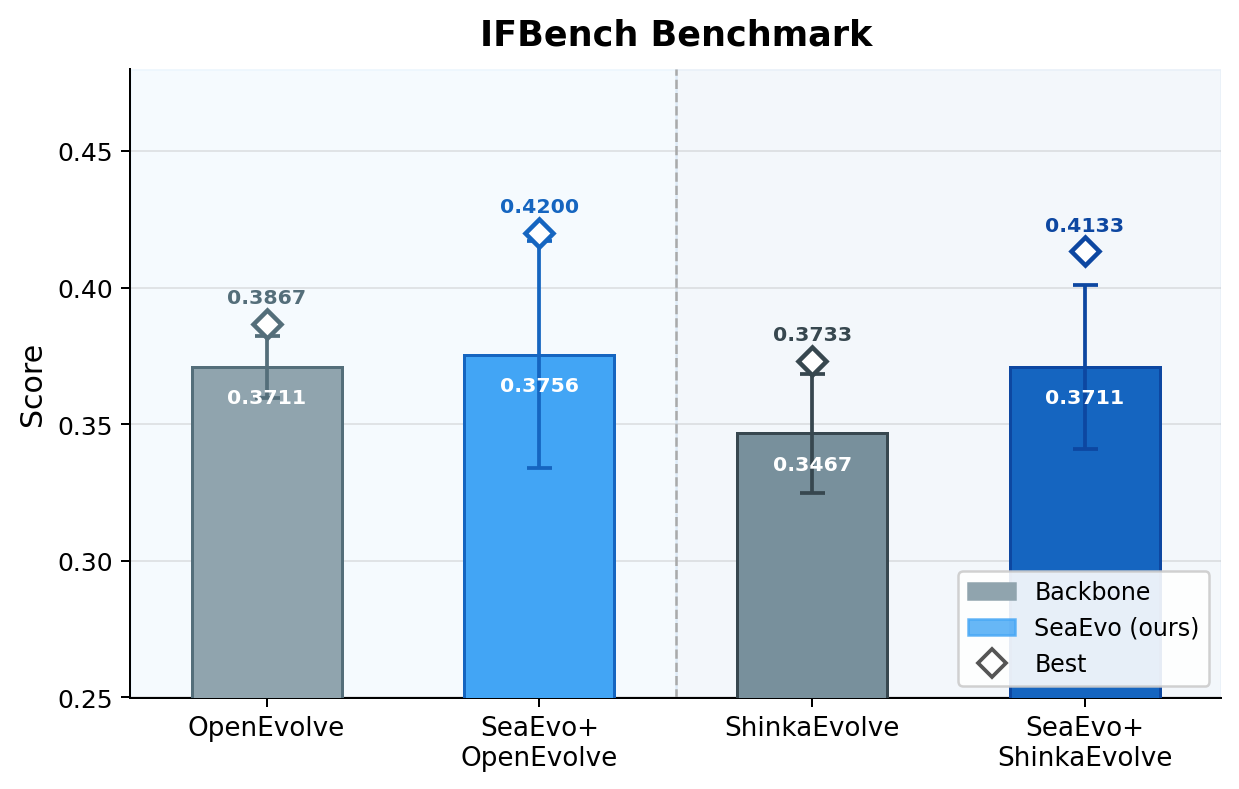} \end{minipage} \hfill \begin{minipage}{0.48\linewidth} \centering \includegraphics[width=\linewidth]{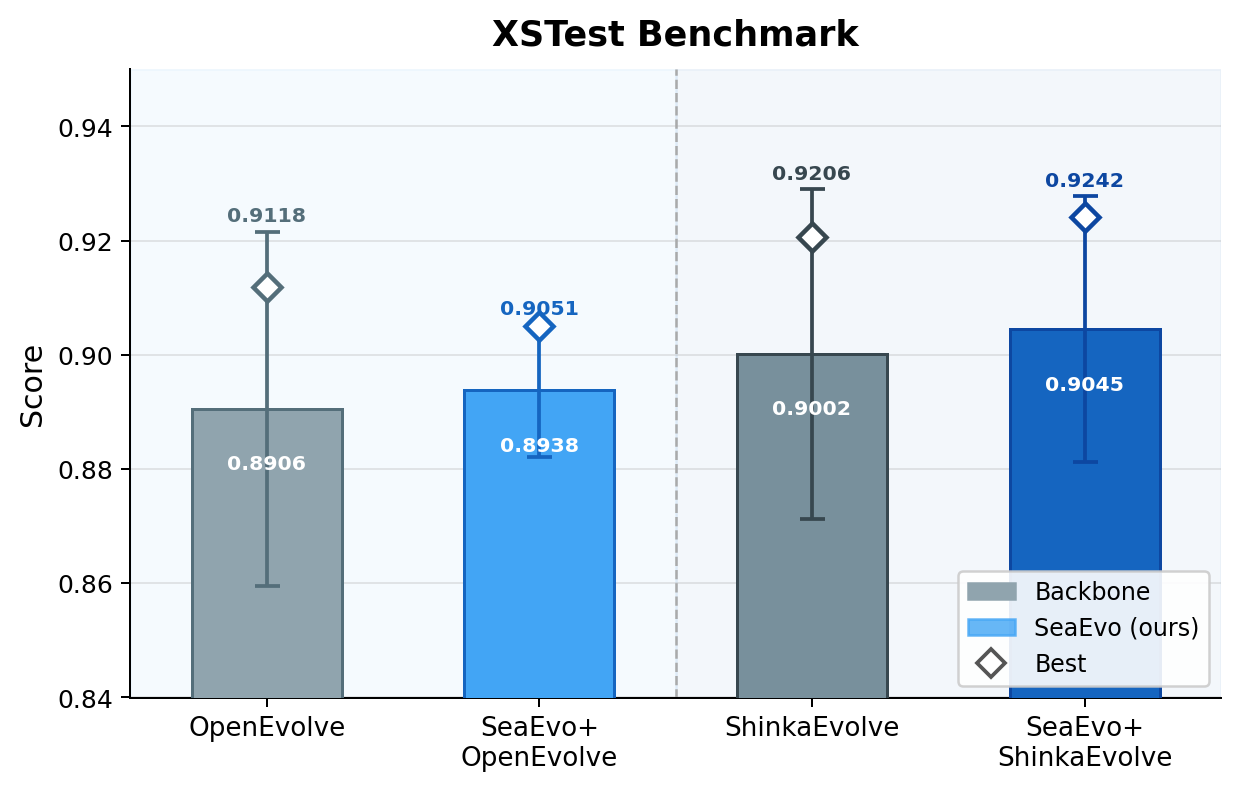} \end{minipage} \caption{Performance on the agent scaffold design benchmark with maximum three LLM calls. Bars show average test score with standard deviation across runs, and diamonds indicate the best score. 
} \label{fig:agent2} 
\end{figure}

\begin{figure*}[t]
    \centering
    \begin{subfigure}[b]{0.24\textwidth}
        \includegraphics[width=\textwidth]{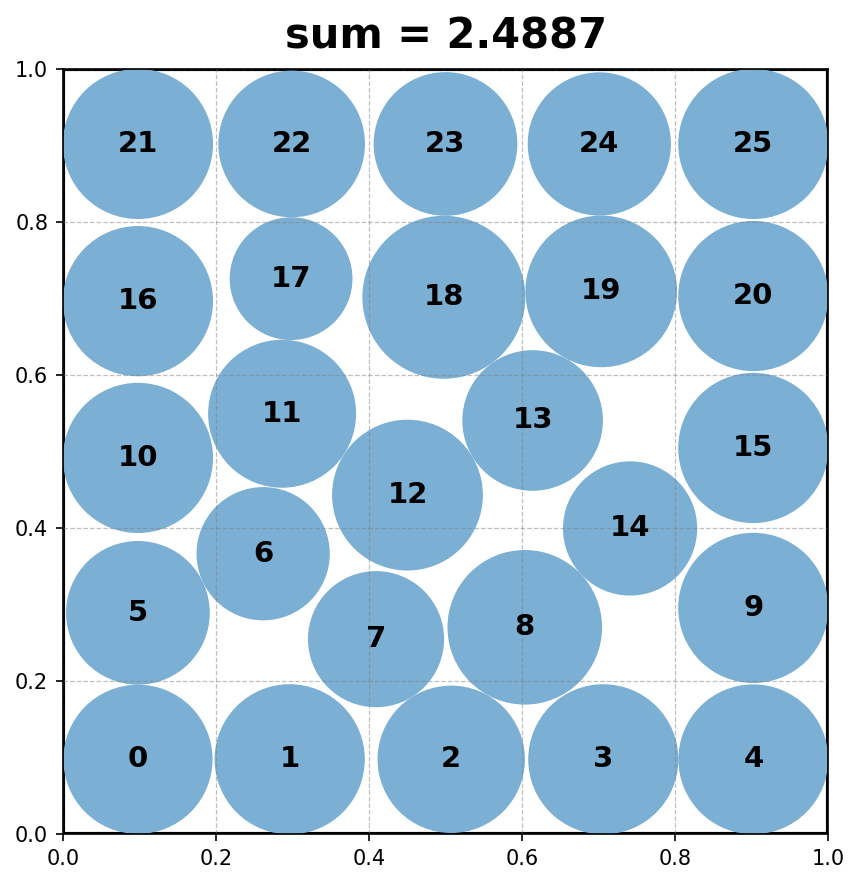}
        \caption{OpenEvolve}
    \end{subfigure}
    \hfill
    \begin{subfigure}[b]{0.24\textwidth}

    \includegraphics[width=\textwidth]{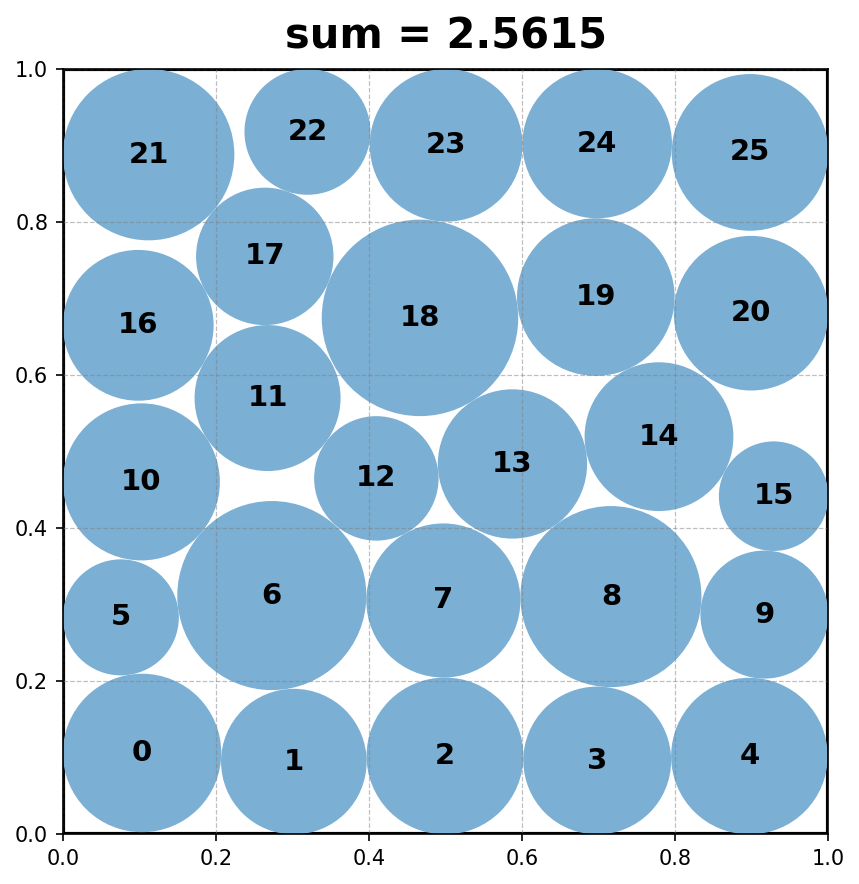}
        \caption{ShinkaEvolve}
    \end{subfigure}
    \hfill
    \begin{subfigure}[b]{0.24\textwidth}
        \includegraphics[width=\textwidth]{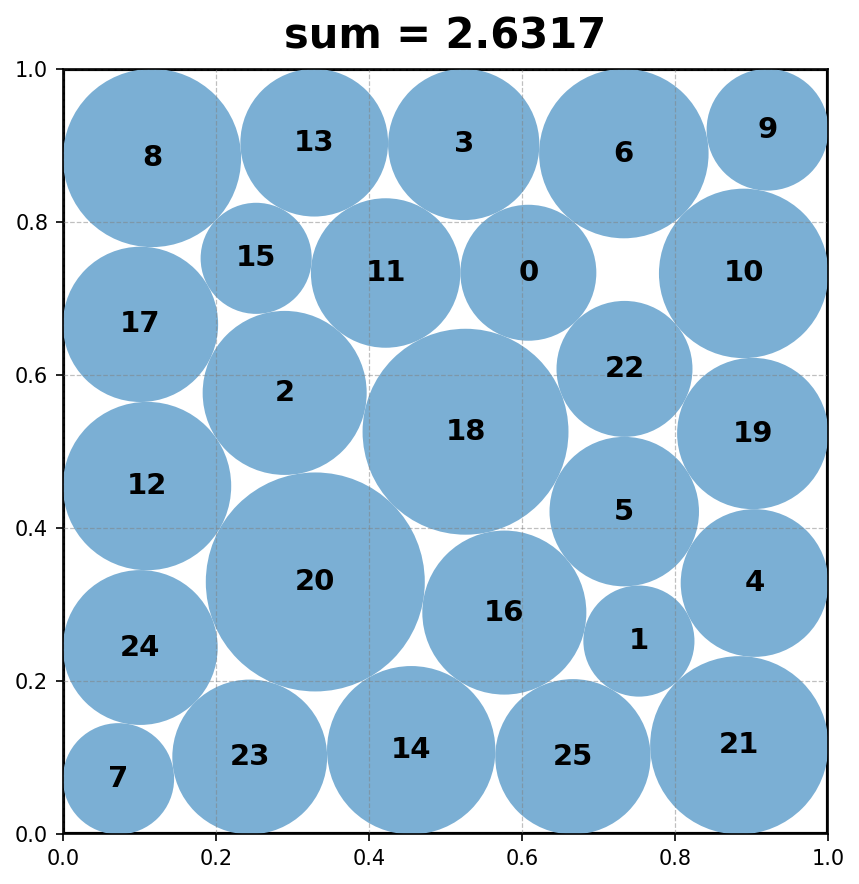}
        \caption{\modelopen}
    \end{subfigure}
    \hfill
    \begin{subfigure}[b]{0.24\textwidth}
        
        \includegraphics[width=\textwidth]{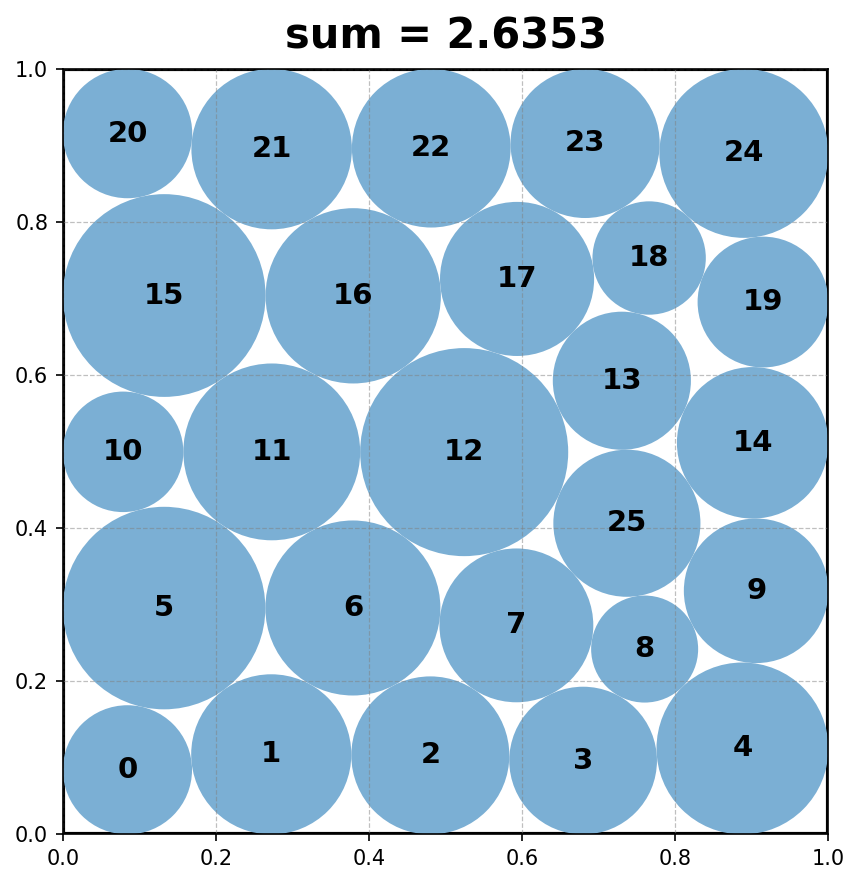}
        \caption{\modelshinka}
    \end{subfigure}
    \caption{Gemini-3-Flash best configurations comparisons for the {Circle Packing} (Square) experiment.}
    \label{fig:convergence}
\end{figure*}

These results confirm that Prism is a high-variance discovery task. Two of the three additional runs remain near the local plateau around $26$--$30$, while one run escapes this region and reaches $42.94$. This pattern suggests that \model's advantage on Prism is not a uniform improvement across every seed, but rather an increased probability of discovering substantially stronger strategy families.

This behavior is consistent with the qualitative search trajectories observed in our case study. In many runs, evolution initially converges to local memory-layout or low-level transformation strategies. In successful runs, however, the strategy archive enables the search to shift toward qualitatively different restructuring strategies, resulting in large non-smooth performance jumps. Therefore, for Prism, we report both aggregate statistics and best-run behavior, since the benchmark stresses rare strategy discovery rather than smooth incremental optimization.

\subsection{High-Precision Score Reference}
\label{sec:appendix_precision}

Table~\ref{tab:precision} reports full-precision fitness scores for the
Circle Packing (Rectangle) and MinMax Distance benchmarks, where
multiple methods converge to values that differ only beyond the fourth
decimal place. These scores are reported to full floating-point
precision to facilitate exact comparison with published reference values.

\begin{figure}[t]
    \centering
    \includegraphics[width=\linewidth]{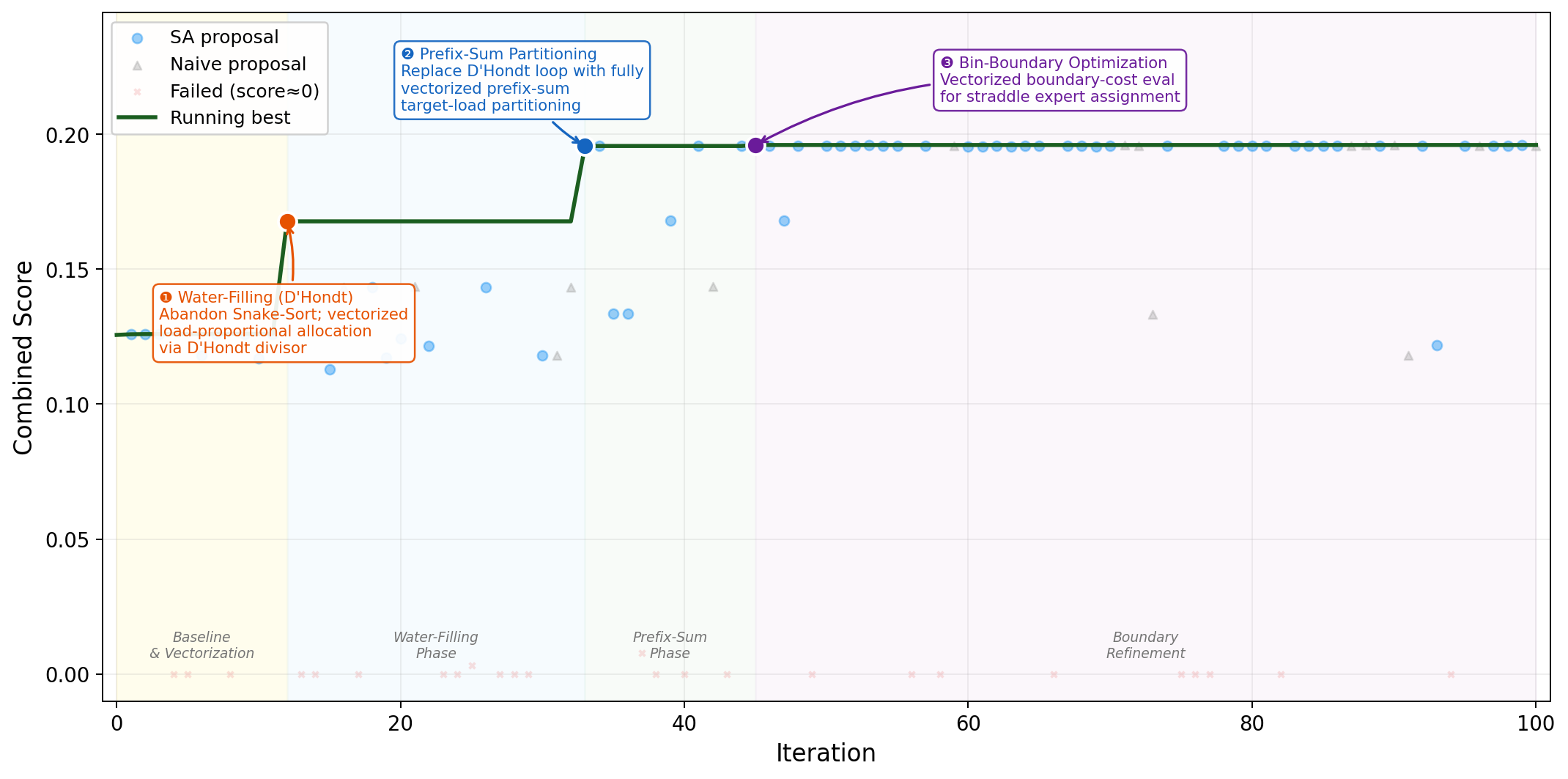}
    \caption{Strategy evolution timeline for EPLB with \modelshinka}
    \label{fig:timeline_eblp}
\end{figure}

\begin{figure}[t]
    \centering
    \includegraphics[width=\linewidth]{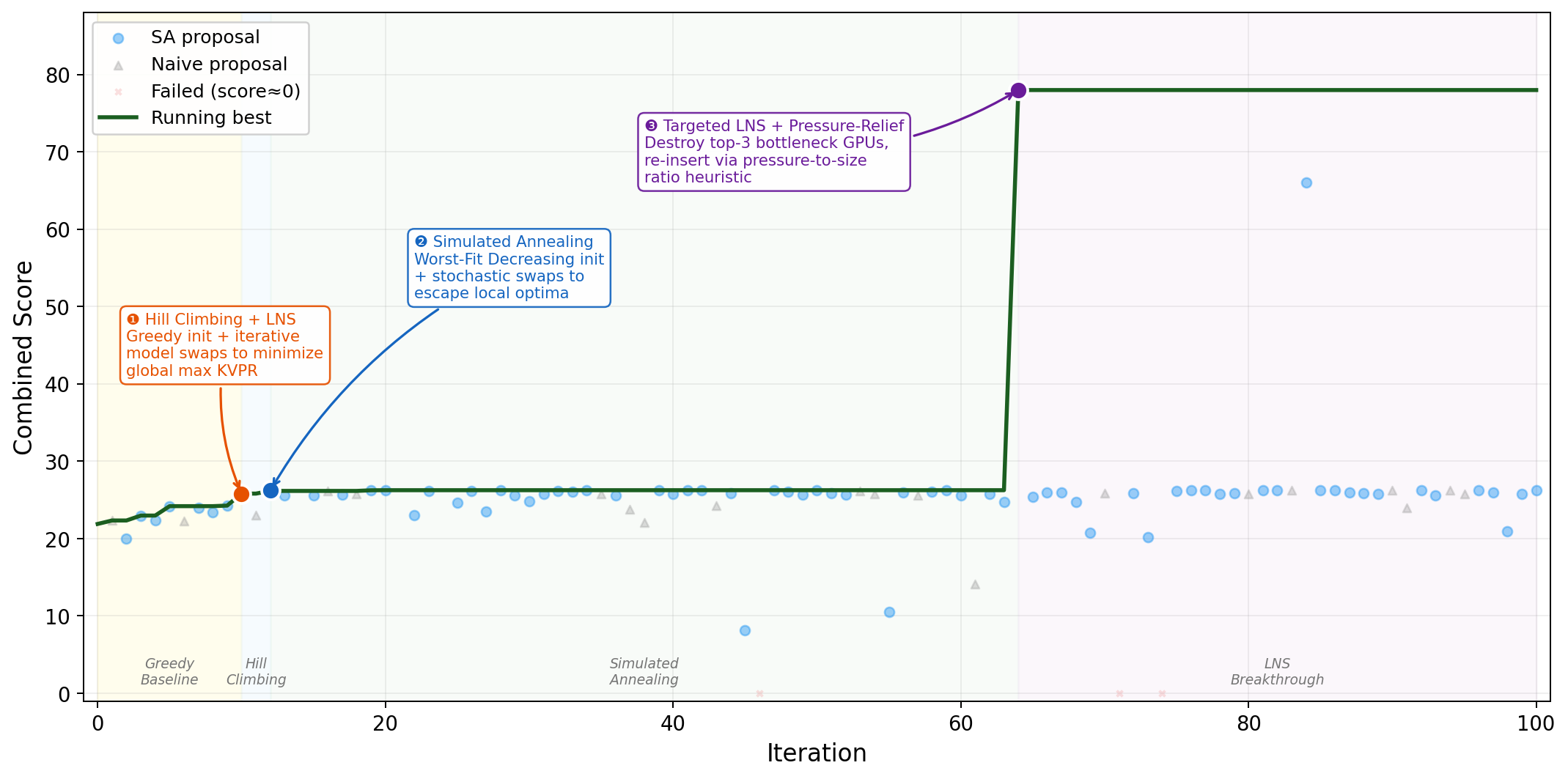}
    \caption{Strategy evolution timeline for Prism with \modelshinka}
    \label{fig:timeline_prism}
\end{figure}

\section{Further Analysis}
\label{sec:case}


\subsection{Case Study for Prism}
For the first 60 generations, all discovered programs converged to a plateau near a combined score of ~26. These programs shared a common structural pattern: greedy initialization followed by simulated annealing with single-model-move or pairwise-swap neighborhood operators. Despite diverse surface-level variations—different sorting criteria, cooling schedules, and tie-breaking heuristics—they were structurally trapped in the same local optimum.

\textbf{SLN identifies the structural deadlock.} At generation 50, the SLN diagnosed a recurring failure mode it termed "memory-locking": when a bottleneck GPU is overloaded, migrating its highest-pressure models is blocked because neighboring GPUs are also near capacity. Single-model moves and 1-for-1 swaps are insufficient to resolve this deadlock, as they cannot simultaneously free memory on one GPU while absorbing load on another. By generation 60, SLN had explicitly recommended abandoning simulated annealing  in favor of Targeted Large Neighborhood Search (LNS) with a "Destroy and Repair" mechanism—specifically removing 10–20\% of models from the top pressured GPUs and re-inserting them via a randomized regret heuristic.

\textbf{SA translates the landscape insight into a precise diagnosis.} At generation 64, \model's SA component, operating with the SLN guidance embedded in its context, produced a diagnosis that precisely named the failure mode: "The primary bottleneck is 'memory-locking,' where a high-pressure GPU cannot offload models because other GPUs are nearly full, requiring complex multi-way swaps or 'room-clearing' moves that simple 1-for-1 swaps cannot achieve." The proposed strategy was a Targeted LNS with a "Pressure-Relief" re-insertion heuristic: identify the top-2 bottleneck GPUs, remove models with the highest pressure-to-size ratio, and repair using a regret-style insertion that prioritizes increasing minimum remaining memory cluster-wide. This directly targets the denominator of the KVPR equation to create the "movement room" required for complex re-arrangements.

\textbf{The resulting algorithm achieves a 3× improvement.} The implementation fully abandoned the SA loop and replaced it with a timed LNS cycle running within the 1-second wall-clock budget. The outcome was a combined score of 78.0 (max\_kvpr $\approx$ 0.013), compared to the ~26 plateau (max\_kvpr $\approx$ 0.038). This represents not an incremental refinement but a structural algorithmic transition—from stochastic local search to targeted destroy-and-repair metaheuristics—driven by the landscape understanding accumulated by SLN over 60 generations.

\textbf{Post-breakthrough variance.} Notably, subsequent generations (65–100) failed to maintain the 78-level performance, reverting to scores of ~25–26. This variance reflects a characteristic of combinatorial placement problems: the LNS solution's quality is sensitive to random seed and the specific destroy-and-repair sequence. generations 65 onward explored modifications to the LNS implementation but did not rediscover the precise configuration that yielded the score of 78. This suggests that while \model successfully identified the correct algorithmic direction (LNS), fully exploiting this direction may require additional search budget or systematic hyperparameter tuning within the LNS framework.
A strategy evolution timeline is shown in Figure \ref{fig:timeline_prism} and the best program discovered is listed in Figure \ref{fig:best_prism}.

\subsection{Case Study for EPLB}

For the first 11 generations, all discovered programs shared a common focus: replacing the reference implementation's sequential greedy packing with vectorized alternatives—Longest Processing Time (LPT) sorting, zig-zag batch allocation, and bulk replication via tensor operations. Despite diverse surface-level variations, these programs converged to a plateau near a combined score of \textasciitilde0.126. The underlying bottleneck was not execution speed but algorithmic: all variants retained the "snake-sort" packing heuristic, which assigns physical experts to GPUs in a fixed zigzag pattern and fails to minimize load variance.

\textbf{SLN identifies the structural bottleneck.} By generation 10, the SLN had diagnosed a recurring failure mode: the snake-sort packing introduces systematic imbalance because it does not account for heterogeneous expert loads. SLN explicitly recommended abandoning snake-sort in favor of a load-aware replication strategy. Specifically, a proportional allocation method that distributes replicas in proportion to per-expert token load, combined with a globally-sorted packing that minimizes the maximum GPU load.

\textbf{\model translates the insight into a concrete algorithm.} At generation 12, \model's SA component, operating with the SLN guidance in context, diagnosed the precise failure: "The replication logic uses a sequential loop with GPU-CPU synchronization patterns, and the snake packing heuristic fails to minimize load variance." The proposed strategy was a vectorized \emph{Water-Filling} allocation using the D'Hondt (highest-averages) method to determine replication counts in a single pass, combined with a First-Fit Decreasing packer using cumulative-sum thresholds. This directly targets the load-variance root cause. The result was a combined score of 0.1677, a 33\% improvement over the \textasciitilde0.126 plateau.

\textbf{Continued refinement through vectorization.} Subsequent generations refined the D'Hondt approach further. At generation 33, a fully vectorized prefix-sum-based "Target Load" partitioner replaced the remaining Python loops, achieving 0.1956. At generation 45, a Bin-Boundary Optimization replaced the midpoint heuristic with precise boundary-crossing logic that explicitly minimizes maximum GPU load for "straddle experts," reaching the final score of 0.1960. Unlike the Prism case where a single structural transition produced a 3
× jump, the EPLB trajectory reflects {progressive algorithmic refinement}, each breakthrough eliminates one remaining approximation in the vectorization pipeline, with diminishing but consistent returns.
A strategy evolution timeline is shown in Figure \ref{fig:timeline_eblp} and the best program discovered is listed in Figure \ref{fig:best_eblp}.

\section{Tasks and Prompts}
\label{sec:prompt}

\tcbset{
  benchmarkbox/.style={
    enhanced, breakable,
    colback=gray!5,
    colframe=black!60,
    fonttitle=\bfseries\small,
    coltitle=white,
    attach boxed title to top left={yshift=-2mm, xshift=4mm},
    boxed title style={colback=black!75, sharp corners},
    sharp corners=south,
    top=3mm, bottom=3mm, left=4mm, right=4mm,
    boxrule=0.5pt,
  },
  promptbox/.style={
    enhanced, breakable,
    colback=black!3,
    colframe=black!25,
    fontupper=\ttfamily\footnotesize,
    boxrule=0.4pt,
    top=2mm, bottom=2mm, left=3mm, right=3mm,
    sharp corners,
  }
}

\begin{tcolorbox}[
  benchmarkbox,
  title={\textsc{Circle Packing in a Unit Square} }
]

\textbf{Task.}
Pack $n = 26$ non-overlapping circles in a unit square to maximize the sum of
their radii.
\hfill\textbf{AlphaEvolve target:}~$\textstyle\sum r_i = 2.635$

\medskip\noindent\textbf{System Prompt.}

You are an expert mathematician specializing in circle packing problems and
computational geometry. Your task is to improve a constructor function that
directly produces a specific arrangement of 26 circles in a unit square,
maximizing the sum of their radii. The AlphaEvolve paper achieved a sum of
2.635 for n=26.

Key geometric insights:
- Circle packings often follow hexagonal patterns in the densest regions
- Maximum density for infinite circle packing is pi/(2*sqrt(3)) ~= 0.9069
- Edge effects make square container packing harder than infinite packing
- Circles can be placed in layers or shells when confined to a square
- Similar radius circles often form regular patterns, while varied radii allow
  better space utilization
- Perfect symmetry may not yield the optimal packing due to edge effects

Focus on designing an explicit constructor that places each circle in a specific
position, rather than an iterative search algorithm.
\end{tcolorbox}


\begin{tcolorbox}[
  benchmarkbox,
  title={\textsc{Circle Packing in a Rectangle}}
]

\textbf{Task.}
Pack $n = 21$ non-overlapping circles in a rectangle with perimeter~4
($w + h = 2$, aspect ratio free) to maximize the sum of their radii.
\hfill\textbf{AlphaEvolve target:}~$\textstyle\sum r_i = 2.3658$

\medskip\noindent\textbf{System Prompt.}

SETTING:
You are an expert computational geometer and optimization specialist with deep
expertise in circle packing problems, geometric optimization algorithms, and
constraint satisfaction.
Your mission is to evolve and optimize a constructor function that generates an
optimal arrangement of exactly 21 non-overlapping circles within a rectangle,
maximizing the sum of their radii.

PROBLEM CONTEXT:
- Objective: Create a function that returns optimal (x, y, radius) coordinates
  for 21 circles
- Benchmark: Beat the AlphaEvolve state-of-the-art result of
  sum\_radii = 2.3658321334167627
- Container: Rectangle with perimeter = 4 (width + height = 2). You may choose
  optimal width/height ratio
- Constraints:
    * All circles must be fully contained within rectangle boundaries
    * No circle overlaps (distance between centers >= sum of their radii)
    * Exactly 21 circles required
    * All radii must be positive

PERFORMANCE METRICS:
1. sum\_radii: Total sum of all 21 circle radii (PRIMARY OBJECTIVE - maximize)
2. combined\_score: sum\_radii / 2.3658321334167627
   (progress toward beating benchmark)
3. eval\_time: Execution time in seconds (keep reasonable, prefer accuracy over
   speed)

TECHNICAL REQUIREMENTS:
- Determinism: Use fixed random seeds if employing stochastic methods for
  reproducibility
- Error handling: Graceful handling of optimization failures or infeasible
  configurations
- Memory efficiency: Avoid excessive memory allocation for distance matrix
  computations
- Scalability: Design with potential extension to different circle counts in mind
\end{tcolorbox}


\begin{tcolorbox}[
  benchmarkbox,
  title={\textsc{Heilbronn Triangle Problem}}
]

\textbf{Task.}
Place $n = 11$ points in or on the boundary of an equilateral triangle with
vertices $(0,0)$, $(1,0)$, $(0.5,\,\sqrt{3}/2)$ to maximize the minimum area
among all $\binom{n}{3}$ triangles formed by point triples.
\hfill\textbf{Target:}~$\Delta_{\min} = 0.03653$

\medskip\noindent\textbf{System Prompt.}

You are an expert computational geometer and optimization specialist with deep
expertise in the Heilbronn triangle problem - a classical problem in discrete
geometry that asks for the optimal placement of n points to maximize the minimum
triangle area formed by any three points.

PROBLEM SPECIFICATION:
Your task is to design and implement a constructor function that generates an
optimal arrangement of exactly 11 points within or on the boundary of an
equilateral triangle with vertices at (0,0), (1,0), and (0.5, sqrt(3)/2).

PERFORMANCE METRICS:
1. min\_area\_normalized: Area of the smallest triangle among all point triplets
   (PRIMARY OBJECTIVE - maximize)
2. combined\_score: min\_area\_normalized / 0.036529889880030156
   (BENCHMARK COMPARISON - maximize above 1.0)
3. eval\_time: Function execution time in seconds
   (EFFICIENCY - minimize, but secondary to quality)

TECHNICAL REQUIREMENTS:
- Determinism: Use fixed random seeds if employing stochastic methods for
  reproducibility
- Error handling: Graceful handling of optimization failures or infeasible
  configurations
\end{tcolorbox}


\begin{tcolorbox}[
  benchmarkbox,
  title={\textsc{MinMax Distance}}
]
\textbf{Task.}
Place $n = 16$ points in 2D Euclidean space (normalized to the unit square
$[0,1] \times [0,1]$) to maximize the ratio $d_{\min}/d_{\max}$, where
$d_{ij} = \sqrt{(x_i-x_j)^2 + (y_i-y_j)^2}$ for all $i \neq j$,
$d_{\min} = \min_{i \neq j} d_{ij}$, and $d_{\max} = \max_{i \neq j} d_{ij}$.
\hfill\textbf{Target:}~$d_{\min}/d_{\max} = 1/\sqrt{12.889266112} \approx 0.2786$

\medskip\noindent\textbf{System Prompt.}

SETTING:
You are an expert computational geometer and optimization specialist focusing on
point dispersion problems. Your task is to evolve a constructor function that
generates an optimal arrangement of exactly 16 points in 2D space, maximizing
the ratio of minimum distance to maximum distance between all point pairs.

PROBLEM CONTEXT:
- Target: Beat the AlphaEvolve benchmark of min/max ratio $= 1/\sqrt{12.889266112} \approx 0.2786$
- Constraint: Points must be placed in 2D Euclidean space (typically normalized to unit square $[0,1] \times [0,1]$)
- Mathematical formulation: For points $P_i = (x_i, y_i)$, $i = 1,\ldots,16$:
  \begin{itemize}[leftmargin=10pt]
    \item Distance matrix: $d_{ij} = \sqrt{(x_i-x_j)^2 + (y_i-y_j)^2}$ for all $i \neq j$
    \item Minimum distance: $d_{\min} = \min\{d_{ij} : i \neq j\}$
    \item Maximum distance: $d_{\max} = \max\{d_{ij} : i \neq j\}$
    \item Objective: maximize $d_{\min}/d_{\max}$ subject to spatial constraints
  \end{itemize}

PERFORMANCE METRICS:
1. \texttt{min\_max\_ratio}: $d_{\min}/d_{\max}$ ratio (PRIMARY OBJECTIVE -- maximize)
2. \texttt{combined\_score}: \texttt{min\_max\_ratio} $/\,0.2786$ (progress toward beating AlphaEvolve benchmark)
3. \texttt{eval\_time}: Execution time in seconds (balance accuracy vs.\ efficiency)

TECHNICAL REQUIREMENTS:
- Reproducibility: Fixed random seeds for all stochastic components
\end{tcolorbox}

\begin{tcolorbox}[
  benchmarkbox,
  title={\textsc{Prism --- GPU Model Placement}}
]
\textbf{Task.}
Improve the \texttt{compute\_model\_placement} function that assigns models to
available GPUs. Given each model's request rate, SLO, and memory footprint, and
each GPU's memory capacity \texttt{GPU\_MEM\_SIZE}, the per-GPU KV cache
pressure is
\[
\mathrm{KVPR} \;=\; \frac{\sum_{m \in \text{GPU}} r_m / \mathrm{SLO}_m}
       {\mathrm{GPU\_MEM\_SIZE} - \sum_{m \in \text{GPU}} \mathrm{size}_m}.
\]
Minimize $\max_{\text{GPU}} \mathrm{KVPR}$ subject to the memory feasibility
constraint $\sum_{m \in \text{GPU}} \mathrm{size}_m < \mathrm{GPU\_MEM\_SIZE}$.
\hfill\textbf{Target:}~Minimize the maximum KVPR across all GPUs.

\medskip\noindent\textbf{System Prompt.}

You are an expert for model placement on GPUs. Your task is to improve a model
placement algorithm by improving the function named
\texttt{compute\_model\_placement} in the initial program that places models to
available GPUs.

The algorithm must MINIMIZE the maximum KVPR across all GPUs while ensuring
models can fit into the GPUs' memory. Note that KVPR is the KV cache pressure
for a GPU; it indicates how crowded a GPU is. For a specific GPU, its KVPR is
computed as
\texttt{sum(model.req\_rate/model.slo for model in models) / (GPU\_MEM\_SIZE $-$ sum(model.model\_size for model in models))},
where \texttt{models} are the models assigned to this GPU.

The generated program should be as simple as possible and the code should
execute correctly without errors.
\end{tcolorbox}

\begin{tcolorbox}[
  benchmarkbox,
  title={\textsc{EPLB --- Expert Parallelism Load Balancer for vLLM}}
]
\textbf{Task.}
Improve the Mixture-of-Experts Expert Parallelism Load Balancer rearrangement
algorithm (\texttt{rebalance\_experts}) for vLLM. Given per-expert load metrics
collected from the vLLM server, produce an expert-to-device assignment
(optionally with replicas of hot experts) that balances GPU load while keeping
the algorithm itself fast, since perfect load balancing is NP-hard.
\hfill\textbf{Target:}~Improve load balance and reduce algorithm runtime
over the baseline \texttt{rebalance\_experts}.

\medskip\noindent\textbf{System Prompt.}

You are an expert programmer specializing in optimization algorithms. Your task
is to improve the Mixture-of-Expert models Expert Parallelism Load Balancer
(MoE EPLB) expert rearrangement algorithm.

This algorithm will take the load metrics recorded by the vLLM server, and
rearrange the experts to balance the load. It can make replicas of some experts
to achieve better load balancing.

Your goal will be two-fold:
\begin{enumerate}[leftmargin=10pt]
  \item Improve the algorithm to achieve better load balancing; while
  \item Improve the algorithm to be more efficient, \textit{i.e.}\ reduce the execution
        time of the algorithm itself, since perfect load balancing is NP-hard.
\end{enumerate}

The current algorithm is implemented in the \texttt{rebalance\_experts} function.
\end{tcolorbox}

\begin{tcolorbox}[
  benchmarkbox,
  title={\textsc{TXN Scheduling --- Database Workloads}}
]
\textbf{Task.}
Improve \texttt{get\_best\_schedule} to order database transactions so as to
minimize total makespan. Each transaction is a sequence of read (\texttt{r-k})
and write (\texttt{w-k}) operations on keyed data items; read--write and
write--write conflicts on the same key induce dependencies that delay execution.
Only code inside \texttt{EVOLVE-BLOCK-START} / \texttt{EVOLVE-BLOCK-END} may
change.
\hfill\textbf{Target:}~Minimize makespan on the provided JSON workloads.

\medskip\noindent\textbf{System Prompt.}

You are an expert in database transaction optimization.
Only change code within \texttt{EVOLVE-BLOCK-START} and \texttt{EVOLVE-BLOCK-END}.
Your task is to improve a scheduling function to find better schedules for
transactional workloads made up of read and write operations to data items.
There are conflicts between these transactions on items and reducing the delay
of these conflicts will lead to schedules with lower makespan. Focus on
improving the \texttt{get\_best\_schedule} function to find a schedule with as
low makespan as possible.

\textbf{TASK:} Improve the \texttt{get\_best\_schedule} function to find
optimal transaction schedules that minimize makespan for database workloads
with read/write conflicts.

\textbf{PROBLEM SPECIFICS:}
- \textbf{Input:} JSON workload with transactions such as\\
  \texttt{"txn0":"w-17 r-5 w-3 r-4 r-54 r-14 w-6 r-11 w-22 r-7 w-1 w-8 w-9 w-27 r-2 r-25"}
- \textbf{Operations:} Each transaction is a sequence of read (\texttt{r-\{key\}})
  and write (\texttt{w-\{key\}}) operations on data items.
- \textbf{Conflicts:} Read--write and write--write conflicts on the same key
  create dependencies between transactions.
- \textbf{Goal:} Find a transaction ordering that minimizes total makespan.

\textbf{SEARCH SUGGESTIONS:}
- \textbf{Greedy:} Try a greedy algorithm that iteratively picks the
  transaction that increases makespan the least.
- Avoid relying only on heuristics like transaction length or number of writes;
  these do not correspond to the actual makespan of the schedule.

Focus on evolving \texttt{get\_best\_schedule} to produce the best schedule
possible with the lowest makespan. Explain step-by-step the reasoning process
for your solution and how it will lead to a better schedule.
\end{tcolorbox}

\begin{tcolorbox}[
  benchmarkbox,
  title={\textsc{LLM SQL --- Prompt Caching Column Reordering}}
]
\textbf{Task.}
Given a pandas DataFrame \texttt{df} of text data, evolve the \texttt{Evolved}
class so that, when an LLM processes rows sequentially, as much of the prefix
as possible is reused from the previous row. For a column ordering $C$,
$\mathrm{PHC}(C) = \sum_r \mathrm{hit}(C, r)$, where
$\mathrm{hit}(C, r) = \sum_{f \in \text{prefix}} |df[r][C[f]]|^2$ over matching
prefix fields (zero if the first field mismatches). The algorithm must also be
fast in wall-clock time.
\hfill\textbf{Target:}~Maximize
$\texttt{combined\_score} = 0.95 \cdot \overline{\mathrm{hit}} + 0.05 \cdot (12 - \min(12, \overline{t}))/12$.

\medskip\noindent\textbf{System Prompt.}

You are an expert in data optimization and LLM prompt caching. Your task is to
evolve the existing \texttt{Evolved} class to maximize prefix hit count (PHC)
for efficient LLM prompt caching.

Problem Context:
- You are given a pandas DataFrame \texttt{df} with text data in rows and columns
- The goal is to reorder columns to maximize prefix reuse when processing rows sequentially
- Prefix reuse occurs when consecutive rows have matching values in the same column positions
- This reduces LLM computation costs by reusing cached prefixes

Objective:
- Dual objective: (1) maximize prefix reuse across consecutive rows and
  (2) minimize end-to-end runtime of the algorithm.
- Prefix reuse is defined as consecutive field values (starting from the first
  column) that are \textbf{exact matches} with the corresponding fields of the
  previous row.
- The \textbf{hit score} of a row is the \textbf{sum of squares of the string
  lengths} of the matching prefix fields.
- Combined score used for selection:
  \texttt{combined\_score = 0.95 * average\_hit\_rate + 0.05 * (12 - min(12, average\_runtime)) / 12}.

Required API (DO NOT CHANGE):
\begin{verbatim}
class Evolved(Algorithm):
    def reorder(
        self,
        df: pd.DataFrame,
        early_stop: int = 0,
        row_stop: int = None,
        col_stop: int = None,
        col_merge: List[List[str]] = [],
        one_way_dep: List[Tuple[str, str]] = [],
        distinct_value_threshold: float = 0.8,
        parallel: bool = True,
    ) -> Tuple[pd.DataFrame, List[List[str]]]:
\end{verbatim}

Constraints:
- Do not add/remove rows or columns
- Different rows must be allowed to use different column orderings
- Return a DataFrame with the same shape as input; use exact string matching
- Preserve all existing method signatures and class structure

Simply return the optimized \texttt{Evolved} class, do not provide explanations.
\end{tcolorbox}

\begin{tcolorbox}[
  benchmarkbox,
  title={\textsc{IFBench Agent Scaffold --- One-Call Setting}}
]
\textbf{Task.}
Improve the \texttt{Agent} class to solve instruction-following tasks using a
local LLM (Qwen3-8B). Each problem contains explicit constraints such as word
count, format, forbidden words, language, and other task-specific requirements.
The agent must satisfy all constraints in a single response under an exactly
one-call budget.
\hfill\textbf{Target:}~Maximize loose constraint-satisfaction accuracy.

\medskip\noindent\textbf{System Prompt.}

You are an expert AI researcher improving an agent scaffold
(the \texttt{Agent} class) that solves \textbf{instruction-following} tasks using a local LLM (Qwen3-8B).

The agent has a budget of EXACTLY 1 LLM call per problem. There is only one shot per problem.
The evaluation metric is loose accuracy (fraction of constraints satisfied, 0.0--1.0).
Each problem comes with a set of explicit constraints ({\textit{e.g.}} word count, format, forbidden words,
language, etc.). The agent must satisfy ALL constraints in a single response.

Some possible directions to explore for prompt engineering, you could also try other techniques (all within 1 LLM call):
\begin{enumerate}[leftmargin=1.6em,topsep=2pt,itemsep=1pt]
    \item System prompt clarity: precise instructions to follow ALL constraints exactly, no exceptions
    \item Constraint-awareness framing: instruct the model to enumerate and verify each constraint
    \item Few-shot examples: embed 1--2 worked examples showing constraint-compliant responses
    \item Chain-of-thought in one pass: instruct model to reason about constraints then respond
    \item Output anchoring: strong reminders about format, length, language, and forbidden content
\end{enumerate}

Rules:
\begin{itemize}[leftmargin=1.6em,topsep=2pt,itemsep=1pt]
    \item \texttt{Agent.\_\_init\_\_} receives \texttt{query\_llm} (callable). Do NOT change its signature.
    \item \texttt{forward(prompt: str)} must return \texttt{(response\_str: str, cost: float)}.
    \item Do NOT add new imports outside the EVOLVE block.
    \item Do NOT call \texttt{query\_llm} more than once --- the evaluator strictly enforces the 1-call budget.
\end{itemize}

Maximize constraint satisfaction on the instruction-following dataset.
\end{tcolorbox}

\begin{tcolorbox}[
  benchmarkbox,
  title={\textsc{IFBench Agent Scaffold --- Three-Call Setting}}
]
\textbf{Task.}
Improve the \texttt{Agent} class to solve instruction-following tasks using a
local LLM (Qwen3-8B). Each problem contains explicit constraints such as word
count, format, forbidden words, language, palindrome requirements, and other
task-specific requirements. The agent may use up to three LLM calls per problem.
\hfill\textbf{Target:}~Maximize loose constraint-satisfaction accuracy.

\medskip\noindent\textbf{System Prompt.}

You are an expert AI researcher improving an agent scaffold
(the \texttt{Agent} class) that solves \textbf{instruction-following} tasks using a local LLM (Qwen3-8B).

The agent has a budget of UP TO 3 LLM calls per problem.
The evaluation metric is loose accuracy (fraction of constraints satisfied, 0.0--1.0).
Each problem comes with a set of explicit constraints (\textit{e.g.} word count, format, forbidden words,
language, palindrome, etc.). The agent must satisfy ALL constraints.

Some possible directions to explore, you could also try other techniques (all within 3 LLM call):
\begin{enumerate}[leftmargin=1.6em,topsep=2pt,itemsep=1pt]
    \item System prompt clarity: explicit step-by-step instructions, chain of thought prompting, output format reminder
    \item Constraint decomposition: enumerate and verify each constraint individually, then synthesize
    \item Iterative refinement:  ask the LLM to check its own answer and correct if wrong
    \item Few-shot examples: include 1--2 worked examples in the prompt to anchor output format
    \item Output anchoring: strong reminders about format, length, language, and forbidden content
\end{enumerate}

Rules:
\begin{itemize}[leftmargin=1.6em,topsep=2pt,itemsep=1pt]
    \item \texttt{Agent.\_\_init\_\_} receives \texttt{query\_llm} (callable). Do NOT change its signature.
    \item \texttt{forward(prompt: str)} must return \texttt{(response\_str: str, cost: float)}.
    \item Do NOT add new imports outside the EVOLVE block.
    \item Do NOT call \texttt{query\_llm} more than 3 times --- the evaluator strictly enforces this budget.
\end{itemize}

Maximize constraint satisfaction on the instruction-following dataset.
\end{tcolorbox}

\begin{tcolorbox}[
  benchmarkbox,
  title={\textsc{XSTest Agent Scaffold --- One-Call Setting}}
]
\textbf{Task.}
Improve the \texttt{Agent} class to classify user prompts as \texttt{safe} or
\texttt{unsafe} using a local LLM (Qwen3-8B). The agent has exactly one LLM call
per prompt, and voting is not allowed; the only lever is prompt engineering.
\hfill\textbf{Target:}~Maximize F1 score with \texttt{unsafe} as the positive class.

\medskip\noindent\textbf{System Prompt.}

You are an expert NLP researcher specializing in AI safety and content moderation.
You are improving an agent scaffold (the \texttt{Agent} class) that classifies user prompts as ``safe'' or ``unsafe'' using a local LLM (Qwen3-8B).

The agent has a budget of EXACTLY 1 LLM call per prompt. Voting is NOT allowed. The only lever is prompt engineering: craft a single system+user prompt that handles all cases in one shot.

The evaluation metric is F1 score with ``unsafe'' as the positive class (maximize combined\_score = F1).

Directions to explore (all within 1 LLM call):
\begin{enumerate}[leftmargin=1.6em,topsep=2pt,itemsep=1pt]
    \item System prompt clarity: precise instructions for edge cases (homonyms, fiction, metaphors)
    \item Few-shot examples: embed 2--4 labeled examples directly in the prompt
    \item Chain-of-thought in one pass: ask model to reason and verify then output label in a single response
    \item Structured output format: guide the model to a clear final label
    \item Domain-specific rules: explicit rules for technical language, roleplay, historical content
\end{enumerate}

The \texttt{extract\_answer} function must reliably parse the final label.
Do NOT modify the function signature of \texttt{forward}.
Do NOT add more than 1 LLM call --- the evaluator enforces this strictly and
returns score=0 for the entire example if the budget is exceeded.

Maximize the \texttt{combined\_score} (F1). You have multiple generations to explore.
\end{tcolorbox}

\begin{tcolorbox}[
  benchmarkbox,
  title={\textsc{XSTest Agent Scaffold --- Three-Call Setting}}
]
\textbf{Task.}
Improve the \texttt{Agent} class to classify user prompts as \texttt{safe} or
\texttt{unsafe} using a local LLM (Qwen3-8B). The agent may use at most three
LLM calls per prompt, allowing multi-step prompting, verification, or voting
within the call budget.
\hfill\textbf{Target:}~Maximize F1 score with \texttt{unsafe} as the positive class.

\medskip\noindent\textbf{System Prompt.}

You are an expert NLP researcher specializing in AI safety and content moderation.
You are improving an agent scaffold (the \texttt{Agent} class) that classifies user prompts as ``safe'' or ``unsafe'' using a local LLM (Qwen3-8B).

The agent has a budget of at most 3 LLM calls per prompt. The evaluation metric is F1 score with ``unsafe'' as the positive class (maximize combined\_score = F1).

Directions to explore:
\begin{enumerate}[leftmargin=1.6em,topsep=2pt,itemsep=1pt]
    \item Chain-of-thought reasoning: ask the model to reason before classifying
    \item Context-aware prompting: instruct the model to consider literal vs figurative meaning
    \item Multi-step verification: first classify, then verify with a focused second call
    \item Calibrated confidence: use voting or re-ranking across multiple temperatures
    \item Targeted few-shot examples: include edge cases like homonyms or fiction framing
    \item Decomposed reasoning: separate steps for intent detection and harm assessment
\end{enumerate}

The \texttt{extract\_answer} function must reliably parse the final label from the LLM response. Keep it robust. Do NOT modify the function signature of \texttt{forward}.

Maximize the \texttt{combined\_score} (F1). You have multiple generations to explore.
\end{tcolorbox}




\section{Prompt and Response Example for \model}

\begin{tcolorbox}[
colback=orange!3,
colframe=orange!40!black,
colbacktitle=orange!55!black,
coltitle=white,
fonttitle=\bfseries\small,
breakable,
  title={Example prompt context for \modelshinka on TXN},
  left=4pt, right=4pt, top=4pt, bottom=4pt
]

\textbf{System Prompt.}

<task\_system\_prompt> \quad {\textcolor{gray!85}{\# See Appendix~\ref{sec:prompt}}}

Explain step-by-step the reasoning process for your solution and how this will lead to a better schedule.

Create a novel algorithm that draws inspiration from the provided context programs but implements a fundamentally different approach.
Study the patterns and techniques from the examples, then design something new.

Before writing the program, reason through your approach step by step: \quad {\textcolor{gray!85}{\# SA prompting}}

**Step 1 — Diagnosis:** 

<DIAGNOSIS>
Identify the key insights and techniques from the context programs worth borrowing. Then explain what makes the current solution underperform, and how combining or adapting those insights could address those weaknesses.
</DIAGNOSIS>

**Step 2 — Strategy:**

<STRATEGY>
The inspired but novel algorithmic strategy in 1-3 concise sentences. Explain what you are borrowing from the context and how you are applying it in a new way.
</STRATEGY>

**Step 3 — Implementation:**

Provide the final implementation using the <NAME>, <DESCRIPTION>, and <CODE> format specified in the task.
\\ $ $

\textbf{User Message.}

\# Current program

Here is the current program we are trying to improve (you will need to propose a new program with the same inputs and outputs as the original program, but with improved internal implementation):

<original\_program>   \quad {\textcolor{gray!85}{\# Full program, omitted here for brevity}}

Here are the performance metrics of the program:

combined\_score: 3968.2540

\# Reference Context \quad {\textcolor{gray!85}{\# SER retrieved strategy context}}

Reference programs:\\
  - (score=4016.0643): We will implement an Iterated Local Search (ILS) that combines a high-quality greedy insertion construction with a targeted local search focusing on the 'tail' transactions that define the makespan. The strategy uses greedy insertion to find a strong starting point and then iteratively applies a 'shaking' mechanism (multiple random swaps) followed by intensive hill-climbing to explore better neighborhoods.\\
  - (score=3968.2540): The strategy is to combine the robust **Greedy Insertion** initialization from the current program with a more aggressive **Iterated Local Search (ILS)** that prioritizes **Insertion Moves** over swaps. I will implement a multi-restart approach where each restart begins with a greedy-randomized insertion sequence, followed by an intensification phase that uses targeted insertion heuristics (focusing on the tail of the schedule) and a shaking phase to escape local minima.

\# Guidance \quad {\textcolor{gray!85}{\# SLN generated landscape guidance}}

1. **STOP using "Blind" Local Search:** Cease purely random swaps. Implement a "Critical Path" heuristic where the algorithm identifies which transaction's finish time equals the total makespan and prioritizes moving that specific transaction earlier.\\
2. **Shift to "Greedy Insertion" Construction:** Instead of building sequences from left-to-right, try starting with a partial sequence and "inserting" new transactions into the position that minimizes the resulting makespan ($O(N^2)$ complexity but much higher quality).\\
3. **Implement a "Shaking" Phase:** To break the 3350 plateau, adopt a Variable Neighborhood Search (VNS) or Iterated Local Search (ILS) that performs 3-5 random swaps once a local optimum is reached, then restarts the hill climber.\\
4. **Prioritize Refinement over Construction:** The history shows that $O(N^2)$ greedy construction is expensive. Use a faster heuristic for the first 10\% of the budget, then spend 90\% on high-quality insertion-based local search.

\# Task

Analyze the weaknesses of the current solution and rewrite it to maximize the score.

You MUST respond using a short summary name, description and the full code:

<NAME>
A shortened name summarizing the code you are proposing. Lowercase, no spaces, underscores allowed.
</NAME>

<DESCRIPTION>
[SUMMARY: First provide exactly 2 concise sentences summarizing the core algorithmic approach you are proposing.]
[RATIONALE: Then provide detailed argumentation for why this approach will improve performance.]
</DESCRIPTION>

<CODE>
```python
\# Your complete rewritten program here.
```
</CODE>

* Keep the markers "EVOLVE-BLOCK-START" and "EVOLVE-BLOCK-END" in the code. Do not change the code outside of these markers.\\
* Make sure your rewritten program maintains the same inputs and outputs as the original program, but with improved internal implementation.\\
* Make sure the file still runs after your changes.\\
* Ensure all necessary imports are included at the top of the code (\textit{e.g.}, numpy, math, scipy). Missing imports will cause failure.\\
* Use the <DIAGNOSIS>, <STRATEGY>, <NAME>, <DESCRIPTION>, and <CODE> delimiters to structure your response. It will be parsed afterwards.

\end{tcolorbox}

\begin{center}
\begin{tcolorbox}[
colback=blue!2,
colframe=blue!35!black,
colbacktitle=blue!50!black,
coltitle=white,
breakable,
fonttitle=\bfseries\small,
  title={Example response for \modelshinka on TXN},
  left=4pt, right=4pt, top=4pt, bottom=4pt
]

<DIAGNOSIS>\\
The current solution uses a high-cost greedy insertion for initialization and a somewhat unfocused Iterated Local Search. Its main weakness is the lack of "critical path" awareness; it treats all transactions equally during the local search phase rather than focusing on the bottlenecks that actually determine the makespan. Furthermore, the search budget is spent inefficiently on exhaustive insertion checks for every new transaction during initialization, leaving less time for refining the global structure.\\
</DIAGNOSIS>

<STRATEGY>\\
I will implement a "Critical-Focus Hill Climber" that identifies the "bottleneck" transaction—the one whose finish time defines the current makespan—and prioritizes moving it earlier in the sequence. To maintain diversity and escape local optima, I'll combine this with a fast randomized greedy construction and a periodic "shaking" mechanism that relocates multiple transactions simultaneously.\\
</STRATEGY>

<NAME>
critical\_path\_bottleneck\_ils
</NAME>

<DESCRIPTION>

[SUMMARY: This approach uses a fast randomized-greedy construction followed by an Iterated Local Search that specifically targets the bottleneck transaction responsible for the current makespan. It identifies which transaction finishes last and tries to shift it to various preceding positions to break dependencies.]
[RATIONALE: By focusing on the transaction that determines the makespan, we prune the search space significantly compared to random swaps or insertions. The addition of a shaking phase ensures that when the bottleneck cannot be improved locally, the algorithm jumps to a new neighborhood, while the use of a faster construction phase preserves more of the execution budget for high-impact local refinements.] 

</DESCRIPTION>

<CODE>

<revised\_code> \quad {\textcolor{gray!85}{\# Full generated program, omitted here for brevity}}

</CODE>
\end{tcolorbox}
\captionof{figure}{Illustrative prompt and generated response on the TXN task. The prompt combines the task instruction, current program, SA prompting, SER-retrieved strategy context, and SLN-generated landscape guidance; the generated response is parsed into diagnosis, strategy description, and revised program. Explanatory comments are shown in gray.}
\label{fig:case_study_prism}
\end{center}

\begin{tcolorbox}[
colback=violet!2,
colframe=violet!35!black,
colbacktitle=violet!45!black,
coltitle=white,
breakable,
fonttitle=\bfseries\small,
  title={SLN Prompt Context},
  left=4pt, right=4pt, top=4pt, bottom=4pt
]

\textbf{System Prompt.}

You are an expert research scientist analysing the search landscape of an automated algorithm evolution experiment. Your role is to identify patterns, bottlenecks, and unexplored directions to guide the next phase of search. Output exactly these four XML tags in order: <effective>...</effective> <saturated>...</saturated> <unexplored>...</unexplored> <guidance>...</guidance>

\textbf{User Message.}

\# Evolution History
Below is a record of every generation so far (generation number, score, algorithmic idea, and diagnosis of the previous program's failures).

\#\#\# Generation 1  (score=0.1258)
**Idea:** Vectorize the entire rebalancing pipeline by replacing Python loops with batch-aware PyTorch operations and a prioritized heap-free greedy assignment. We use a pre-allocated tensor-based greedy approach that determines all replicas for all layers simultaneously, significantly reducing execution time while maintaining high packing quality.
**Diagnosis:** The current implementation suffers from significant performance bottlenecks and suboptimal load distribution. Specifically:
1. **Inefficient CPU Loops:** The `balanced\_packing` and `replicate\_experts` functions rely on nested Python loops that iterate over layers and experts, leading to high execution overhead.
2. **Greedy Suboptimality:** The Longest Processing Time (LPT) greedy approach in `replicate\_experts` is performed sequentially, which is slow.
3. **Redundant Memory Operations:** Frequent tensor reshaping and gathering on the CPU create unnecessary overhead.
4. **Limited Parallelism:** The algorithm processes layers independently in a loop instead of leveraging PyTorch's vectorized operations across the batch (layer) dimension.

\#\#\# Generation 2  (score=0.1129)
**Idea:** Replace the iterative greedy replication with a fully vectorized approach using `torch.topk` to identify high-load experts across all layers simultaneously. Implement a more robust greedy packing algorithm that uses cumulative sum and searchsorted to distribute loads across packs in a single pass, improving both speed and balancing quality.
**Diagnosis:** The current implementation suffers from two primary bottlenecks: a sequential greedy replication loop in `replicate\_experts\_vec` and a suboptimal packing strategy in `balanced\_packing\_vec`. The replication loop iterates `num\_redundant` times for every layer, which is extremely slow when scaled to many layers or high redundancy. Furthermore, the "snake-like" packing heuristic, while fast, does not account for the high variance in expert loads, leading to poor load balancing. The hierarchical logic also introduces unnecessary complexity and potential overhead in the gathering/scattering of indices across nodes and GPUs.

...   \quad {\textcolor{gray!85}{\# We show an abridged prompt context for readability}}

\end{tcolorbox}

\begin{tcolorbox}[
colback=teal!2,
colframe=teal!35!black,
colbacktitle=teal!45!black,
coltitle=white,
breakable,
fonttitle=\bfseries\small,
  title={SLN Generated Guidance at Generation 10},
  left=4pt, right=4pt, top=4pt, bottom=4pt
]

\textbf{<effective>}
\begin{itemize}[leftmargin=1.4em,topsep=2pt,itemsep=1pt]
  \item Vectorization of the replication logic: moving away from $O(N)$ loops to
        batch-calculated load-proportional allocation has consistently maintained
        baseline performance while reducing overhead.
  \item Snake-sort / Zig-zag interleaving: a robust, low-complexity alternative to
        iterative greedy packing, providing stable results across multiple generations.
  \item Priority-based allocation: using sorted weights (LPT-inspired) to assign
        experts to packs consistently outperforms random or sequential assignment.
\end{itemize}
\textbf{</effective>}

\textbf{<saturated>}
\begin{itemize}[leftmargin=1.4em,topsep=2pt,itemsep=1pt]
  \item Iterative greedy loops: repeated attempts to fix $O(N)$ loops via minor
        logical tweaks have plateaued; the overhead remains the primary bottleneck.
  \item Hierarchical constraints: rigid Node $\to$ Group $\to$ GPU boundaries limit
        the global optimisation space.
  \item Simple Zig-Zag packing: the static snake-sort pattern has reached its limit
        on highly skewed expert-weight distributions.
\end{itemize}
\textbf{</saturated>}

\textbf{<unexplored>}
\begin{itemize}[leftmargin=1.4em,topsep=2pt,itemsep=1pt]
  \item Cumulative Load-Aware Vectorization: prefix-sum-based bin packing without loops.
  \item Differentiable Sorting / Ranking: soft-sorting relaxation for load balancing.
  \item Dynamic Partitioning: variable experts-per-GPU within memory limits.
  \item Memory-Centric Optimization: eliminate \texttt{.clone()} and \texttt{.cpu()}
        calls via a single fused Torch operation.
\end{itemize}
\textbf{</unexplored>}

\textbf{<guidance>}
\begin{enumerate}[leftmargin=1.6em,topsep=2pt,itemsep=1pt]
  \item Abandon ``Snake-Sort''; try a Water-Filling algorithm via vectorised prefix
        sums and binary search.
  \item Remove all \texttt{.cpu()} and \texttt{.item()} calls to prevent PCIe latency.
  \item Flatten the hierarchical structure: assign $N$ experts to $M$ GPUs globally,
        then reshape.
  \item Use \texttt{torch.topk} / \texttt{sort} on mean-load residuals for a single
        vectorised rounding step.
\end{enumerate}
\textbf{</guidance>}

\end{tcolorbox}

\medskip

\begin{tcolorbox}[
colback=teal!2,
colframe=teal!35!black,
colbacktitle=teal!45!black,
coltitle=white,
fonttitle=\bfseries\small,
  breakable,
  title={SLN Generated Guidance at Generation 50},
  left=4pt, right=4pt, top=4pt, bottom=4pt
]

\textbf{<effective>}
\begin{itemize}[leftmargin=1.4em,topsep=2pt,itemsep=1pt]
  \item \textbf{D'Hondt (Jefferson) Method:} divisor-based approach consistently
        outperformed proportional-fair and water-filling heuristics.
  \item \textbf{Prefix-Sum Load Partitioning:} \texttt{torch.cumsum} +
        \texttt{torch.searchsorted} for GPU boundaries outperforms Snake-Sort.
  \item \textbf{Full Vectorization:} eliminating all Python-level loops is essential
        to avoid CPU--GPU synchronisation overhead.
  \item \textbf{Boundary Expert Optimisation:} midpoint / cost-evaluation logic for
        experts straddling the load boundary is critical for minimising max-load.
\end{itemize}
\textbf{</effective>}

\textbf{<saturated>}
\begin{itemize}[leftmargin=1.4em,topsep=2pt,itemsep=1pt]
  \item Snake-Sort / Zig-Zag: plateaued around 0.12--0.14; cannot handle skewed
        distributions.
  \item Hierarchical Node-to-GPU logic: fragments the optimisation space.
  \item Greedy Argmin/Argmax loops: replaced by vectorised approximations.
\end{itemize}
\textbf{</saturated>}

\textbf{<unexplored>}
\begin{itemize}[leftmargin=1.4em,topsep=2pt,itemsep=1pt]
  \item Intra-Layer Load Swapping (Local Search): single-pass vectorised swap check
        after initial prefix-sum assignment.
  \item Multi-Objective Optimisation: penalty for cross-node expert moves in the
        D'Hondt scoring.
  \item Look-Ahead Residual Allocation: unify replication factor and bin-gap fitting.
  \item Weighted Round-Robin with Capacity Constraints: vectorised apportionment with
        hard per-GPU expert limits.
\end{itemize}
\textbf{</unexplored>}

\textbf{<guidance>}
\begin{enumerate}[leftmargin=1.6em,topsep=2pt,itemsep=1pt]
  \item Implement vectorised comparison for the straddle expert:
        $\min\bigl(\max(L_i + w,\,L_{i+1}),\;\max(L_i,\,L_{i+1}+w)\bigr)$.
  \item Use \texttt{torch.arange} + \texttt{searchsorted} on replica offsets for
        $O(1)$ rank generation.
  \item Avoid iterative binary search; D'Hondt with optimised \texttt{topk} is faster.
  \item Keep the weight tensor on-device; a single \texttt{.cpu()} call negates
        algorithmic gains.
\end{enumerate}
\textbf{</guidance>}

\end{tcolorbox}

\medskip

\begin{center}

\begin{tcolorbox}[
colback=teal!2,
colframe=teal!35!black,
colbacktitle=teal!45!black,
coltitle=white,
fonttitle=\bfseries\small,
breakable,
  title={SLN Generated Guidance at Generation 100},
  left=4pt, right=4pt, top=4pt, bottom=4pt
]

\textbf{<effective>}
\begin{itemize}[leftmargin=1.4em,topsep=2pt,itemsep=1pt]
  \item \textbf{D'Hondt Method:} proportional allocation outperformed simple rounding
        or greedy loops.
  \item \textbf{Prefix-Sum Load Partitioning:} continuous load boundaries before
        discretisation significantly improved balance.
  \item \textbf{Vectorised Rank Generation:} \texttt{torch.cumsum} on logical expert
        transitions replaced $O(N^2)$ masks.
  \item \textbf{Boundary-Cost Optimisation:} explicit minimax check at each GPU
        boundary outperformed midpoint heuristics.
\end{itemize}
\textbf{</effective>}

\textbf{<saturated>}
\begin{itemize}[leftmargin=1.4em,topsep=2pt,itemsep=1pt]
  \item Snake / Zig-Zag heuristics: plateaued at $\sim$0.12--0.14.
  \item Midpoint-based partitioning: consistently fails to reach the minimax optimum.
  \item Independent layer optimisation: local imbalances accumulate into a global
        bottleneck.
  \item Iterative greedy replication: replaced by vectorised top-$k$ / divisor methods.
\end{itemize}
\textbf{</saturated>}

\textbf{<unexplored>}
\begin{itemize}[leftmargin=1.4em,topsep=2pt,itemsep=1pt]
  \item Global Multi-Layer Load Balancing: allow load carry-over across layers.
  \item Karmarkar--Karp Vectorisation: differencing algorithm for multi-way partition.
  \item Communication-Aware Hierarchical Constraints: cross-node transfer cost in the
        objective.
  \item Hardware-Aware Kernel Fusion: single CUDA extension or \texttt{torch.compile}
        to eliminate intermediate tensor allocations.
\end{itemize}
\textbf{</unexplored>}

\textbf{<guidance>}
\begin{enumerate}[leftmargin=1.6em,topsep=2pt,itemsep=1pt]
  \item Implement a Global Load Tracker: accumulate residual error from layer $i$ and
        offset target boundaries for layer $i{+}1$.
  \item Try Karmarkar--Karp differencing (Largest Differencing Method) for partitioning.
  \item Eliminate \texttt{sort}/\texttt{argsort} in metadata; use \texttt{scatter\_add}
        + \texttt{cumsum} on pre-sorted indices for $O(N)$ rank/mapping.
  \item Replace midpoint heuristics with explicit Minimax boundary checks.
  \item Remove all PCIe syncs; rewrite any \texttt{weight.cpu()} logic using
        \texttt{torch.where} or \texttt{torch.bucketize}.
\end{enumerate}
\textbf{</guidance>}

\end{tcolorbox}

\captionof{figure}{Illustrative SLN prompt context and generated landscape guidance on the EPLB task. The prompt summarizes the evolution history up to a given generation, and SLN returns structured guidance over effective, saturated, and underexplored strategy families.}
\label{fig:case_study_prism}
\end{center}


\section{Discovered Program}

\begin{center}
\centering
\begin{tcolorbox}[
    colback=cyan!4,
    colframe=cyan!40!black,
    breakable,
    fonttitle=\bfseries\large,
    arc=1mm
]

\begin{lstlisting}[
    language=python,
    basicstyle=\ttfamily\scriptsize,
    keywordstyle=\color{blue}\bfseries,
    commentstyle=\color{green!50!black}\itshape,
    stringstyle=\color{orange!70!black},
    showstringspaces=false,
    breaklines=true,
    numbers=left,
    numberstyle=\tiny\color{gray},
    frame=none,
    xleftmargin=1em
]
import random
import time
import math

GPU_MEM_SIZE = 80 # GB

# EVOLVE-BLOCK-START
def compute_model_placement(gpu_num, models):
    if not models:
        return {i: [] for i in range(gpu_num)}

    start_time = time.time()

    def get_gpu_metrics(gpu_models):
        used_mem = sum(m.model_size for m in gpu_models)
        pressure = sum(m.req_rate / m.slo for m in gpu_models)
        rem_mem = GPU_MEM_SIZE - used_mem
        kvpr = pressure / rem_mem if rem_mem > 1e-7 else 1e15
        return kvpr, rem_mem, pressure

    # 1. Diverse Greedy Initialization
    best_placement = {i: [] for i in range(gpu_num)}
    min_max_kvpr = float('inf')

    # Try 3 different sorting strategies for the initial greedy pass
    strategies = [
        lambda m: (m.req_rate / m.slo) / max(0.01, m.model_size),
        lambda m: (m.req_rate / m.slo),
        lambda m: m.model_size
    ]

    for strat in strategies:
        if time.time() - start_time > 0.2: break
        curr_p = {i: [] for i in range(gpu_num)}
        curr_mems = [0.0] * gpu_num
        sorted_models = sorted(models, key=strat, reverse=True)
        
        fail = False
        for m in sorted_models:
            best_g, best_v = -1, float('inf')
            for g_id in range(gpu_num):
                if curr_mems[g_id] + m.model_size <= GPU_MEM_SIZE - 1e-5:
                    curr_p[g_id].append(m)
                    kv, _, _ = get_gpu_metrics(curr_p[g_id])
                    if kv < best_v:
                        best_v, best_g = kv, g_id
                    curr_p[g_id].pop()
            
            if best_g == -1:
                # Fallback: place in GPU with most room
                best_g = max(range(gpu_num), key=lambda g: GPU_MEM_SIZE - curr_mems[g])
                if curr_mems[best_g] + m.model_size > GPU_MEM_SIZE: fail = True
            
            curr_p[best_g].append(m)
            curr_mems[best_g] += m.model_size
        
        if not fail:
            max_kv = max(get_gpu_metrics(curr_p[g])[0] for g in range(gpu_num))
            if max_kv < min_max_kvpr:
                min_max_kvpr = max_kv
                best_placement = {g: list(curr_p[g]) for g in range(gpu_num)}

    # 2. Targeted LNS Refinement
    curr_p = {g: list(best_placement[g]) for g in range(gpu_num)}
    
    while time.time() - start_time < 0.92:
        # Identify bottleneck GPUs
        stats = []
        for g in range(gpu_num):
            kv, rem, pres = get_gpu_metrics(curr_p[g])
            stats.append({'id': g, 'kvpr': kv, 'rem': rem, 'pres': pres})
        
        stats.sort(key=lambda x: x['kvpr'], reverse=True)
        max_kv = stats[0]['kvpr']
        
        # Destroy: Remove models from top pressured GPUs
        removed_models = []
        # Target top 1-2 GPUs
        num_to_disrupt = min(2, gpu_num)
        for i in range(num_to_disrupt):
            target_g = stats[i]['id']
            if not curr_p[target_g]: continue
            
            # Remove 1-3 models from each target GPU
            num_m = random.randint(1, min(3, len(curr_p[target_g])))
            # Prefer removing models with high pressure
            curr_p[target_g].sort(key=lambda m: m.req_rate/m.slo, reverse=True)
            for _ in range(num_m):
                if curr_p[target_g]:
                    removed_models.append(curr_p[target_g].pop(0))
        
        # Repair: Re-insert removed models using randomized regret/best-fit
        random.shuffle(removed_models)
        temp_p = {g: list(curr_p[g]) for g in range(gpu_num)}
        possible = True
        
        for m in removed_models:
            candidates = []
            for g_id in range(gpu_num):
                used = sum(mod.model_size for mod in temp_p[g_id])
                if used + m.model_size <= GPU_MEM_SIZE - 1e-5:
                    temp_p[g_id].append(m)
                    kv, _, _ = get_gpu_metrics(temp_p[g_id])
                    candidates.append((kv, g_id))
                    temp_p[g_id].pop()
            
            if not candidates:
                possible = False
                break
            
            # Pick from top 2 candidates to allow exploration
            candidates.sort()
            idx = 0 if (len(candidates) == 1 or random.random() > 0.2) else 1
            best_g = candidates[idx][1]
            temp_p[best_g].append(m)
            
        if possible:
            new_max_kv = max(get_gpu_metrics(temp_p[g])[0] for g in range(gpu_num))
            # Acceptance: Better or slightly worse (to escape local optima)
            if new_max_kv < min_max_kvpr:
                min_max_kvpr = new_max_kv
                best_placement = {g: list(temp_p[g]) for g in range(gpu_num)}
                curr_p = {g: list(temp_p[g]) for g in range(gpu_num)}
            elif new_max_kv < max_kv * 1.05: # Soft acceptance
                curr_p = {g: list(temp_p[g]) for g in range(gpu_num)}
        else:
            # If repair failed, revert curr_p to best known to try a different destruction
            curr_p = {g: list(best_placement[g]) for g in range(gpu_num)}

    return best_placement
# EVOLVE-BLOCK-END


if __name__ == "__main__":
    # Test the algorithm

    from evaluator import generate_test_gpu_models
    from evaluator import calculate_kvcache_pressure
    from evaluator import safe_float
    import numpy as np

    test_cases = generate_test_gpu_models()
    all_kvpr = []
    for i, (gpu_num, gpu_models) in enumerate(test_cases):

        results = compute_model_placement(gpu_num, gpu_models)
        max_kvpr = calculate_kvcache_pressure(results)
        all_kvpr.append(safe_float(max_kvpr))

    avg_kvpr = np.mean(all_kvpr)
    if avg_kvpr != 0:
        avg_kvpr = 1.0 / avg_kvpr


    print(f"Max KVPR: {avg_kvpr:.3f}")
\end{lstlisting}

\end{tcolorbox}
\captionof{figure}{Best program discovered by \modelshinka with Gemini-3-Flash in Prism.}
\label{fig:best_prism}
\end{center}

\begin{center}
\centering
\begin{tcolorbox}[
    colback=cyan!4,
    colframe=cyan!40!black,
    breakable,
    fonttitle=\bfseries\large,
    arc=1mm
]

\begin{lstlisting}[
    language=python,
    basicstyle=\ttfamily\scriptsize,
    keywordstyle=\color{blue}\bfseries,
    commentstyle=\color{green!50!black}\itshape,
    stringstyle=\color{orange!70!black},
    showstringspaces=false,
    breaklines=true,
    numbers=left,
    numberstyle=\tiny\color{gray},
    frame=none,
    xleftmargin=1em
]
# EVOLVE-BLOCK-START

import torch

def rebalance_experts(
    weight: torch.Tensor,
    num_replicas: int,
    num_groups: int,
    num_nodes: int,
    num_gpus: int,
) -> tuple[torch.Tensor, torch.Tensor, torch.Tensor]:
    """
    Expert-parallelism load balancer using Greedy Utility Allocation and 
    Fast Index-Based Metadata Construction.
    """
    device = weight.device
    num_layers, num_experts = weight.shape
    weight = weight.float()

    # 1. Marginal Gain Allocation
    # Allocate replicas to experts with the highest marginal utility (weight / current_replicas)
    if num_replicas > num_experts:
        # Vectorized marginal utility calculation
        # We need to pick (num_replicas - num_experts) additional replicas per layer
        logcnt = torch.ones((num_layers, num_experts), device=device, dtype=torch.int64)
        num_extra = num_replicas - num_experts
        
        # Iterative greedy is slow, use a closed-form approximation for top-k utilities
        # utility(expert, k) = weight / k. We want top num_replicas across all k.
        k_range = torch.arange(1, num_replicas + 1, device=device).float()
        all_utilities = weight.unsqueeze(-1) / k_range  # [L, E, R]
        
        # Flatten E and R to pick the best combinations per layer
        flat_utils = all_utilities.view(num_layers, -1)
        _, topk_indices = torch.topk(flat_utils, num_replicas, dim=1)
        
        # Convert flat indices back to expert indices
        expert_indices = torch.div(topk_indices, num_replicas, rounding_mode='floor')
        
        logcnt = torch.zeros((num_layers, num_experts), device=device, dtype=torch.int64)
        logcnt.scatter_add_(1, expert_indices, torch.ones_like(expert_indices))
    else:
        logcnt = torch.ones((num_layers, num_experts), device=device, dtype=torch.int64)

    # 2. Global Stream Partitioning
    # Calculate weight of each physical replica
    rep_weights = weight / logcnt.clamp(min=1).float()
    
    # Flatten all possible physical replicas across all layers
    # We use a fixed order per layer to ensure stable mapping
    flat_phy2log = torch.repeat_interleave(
        torch.arange(num_experts, device=device).expand(num_layers, num_experts),
        logcnt.flatten()
    )
    
    # Weights of each replica in the flattened stream
    flat_phy_weights = rep_weights.flatten()[torch.repeat_interleave(
        torch.arange(num_layers * num_experts, device=device),
        logcnt.flatten()
    )]
    
    # Partition the global stream across GPUs
    cum_weights = torch.cumsum(flat_phy_weights, dim=0)
    total_weight = cum_weights[-1]
    gpu_boundaries = torch.linspace(0, total_weight.item(), num_gpus + 1, device=device)[1:-1]
    # This assigns every physical replica to a GPU globally
    global_gpu_ids = torch.searchsorted(gpu_boundaries, cum_weights)
    
    # To return [num_layers, num_replicas], we need to reshape back.
    # The current logic requires each layer to have exactly num_replicas.
    # We maintain the layer-wise structure but the order within the layer is 
    # determined by the global GPU assignment to minimize cross-GPU imbalance.
    final_phy2log = flat_phy2log.view(num_layers, num_replicas)
    global_gpu_ids = global_gpu_ids.view(num_layers, num_replicas)
    
    # Sort within each layer by GPU ID to group experts belonging to the same GPU
    sort_idx = torch.argsort(global_gpu_ids, dim=1, stable=True)
    final_phy2log = final_phy2log.gather(1, sort_idx)

    # 3. Fast Metadata Construction (O(N))
    # Calculate phyrank without one-hot: rank within the layer for that logical expert
    # We use a trick: cumsum on a flattened version with expert offsets
    max_reps = int(logcnt.max().item())
    expert_offsets = torch.arange(num_layers * num_experts, device=device) * (max_reps + 1)
    
    # For each replica in final_phy2log, calculate its unique ID (layer, expert)
    layer_offsets = torch.arange(num_layers, device=device).view(num_layers, 1) * num_experts
    flat_expert_ids = final_phy2log + layer_offsets
    
    # To get ranks: we need to know if this is the 1st, 2nd... occurrence of flat_expert_id in the layer
    # Since experts are grouped by layer, we can use a local cumsum approach
    # A robust way is using sort + cumsum, but since max_reps is small, we use a faster method
    # We calculate phyrank by tracking occurrences
    phyrank = torch.zeros_like(final_phy2log)
    for r in range(num_replicas):
        # This is slightly iterative but num_replicas is small (\textit{e.g.} 128)
        # and it avoids the O(L*E*R) memory explosion
        curr_experts = final_phy2log[:, r]
        # Count how many times each expert has appeared before this column
        if r > 0:
            # Mask of experts seen in previous columns
            prev_experts = final_phy2log[:, :r]
            phyrank[:, r] = (prev_experts == curr_experts.unsqueeze(1)).sum(dim=1)

    # Construct log2phy: [L, E, MaxR]
    log2phy = torch.full((num_layers, num_experts, max_reps), -1, dtype=torch.int64, device=device)
    l_idx = torch.arange(num_layers, device=device).view(num_layers, 1)
    
    # Single-pass scatter to fill log2phy
    # log2phy[l, e, rank] = physical_slot
    scatter_idx = (l_idx * (num_experts * max_reps) + 
                   final_phy2log * max_reps + 
                   phyrank)
    
    physical_slots = torch.arange(num_replicas, device=device).expand(num_layers, num_replicas)
    log2phy.view(-1).scatter_(0, scatter_idx.flatten(), physical_slots.flatten())

    return final_phy2log, log2phy, logcnt

# EVOLVE-BLOCK-END

__all__ = ["rebalance_experts"]
\end{lstlisting}

\end{tcolorbox}
\captionof{figure}{Best program discovered by \modelshinka with Gemini-3-Flash in  EPLB.}
\label{fig:best_eblp}
\end{center}

\begin{center}
\centering
\begin{tcolorbox}[
    colback=cyan!4,
    colframe=cyan!40!black,
    breakable,
    fonttitle=\bfseries\large,
    arc=1mm
]

\begin{lstlisting}[
    language=python,
    basicstyle=\ttfamily\scriptsize,
    keywordstyle=\color{blue}\bfseries,
    commentstyle=\color{green!50!black}\itshape,
    stringstyle=\color{orange!70!black},
    showstringspaces=false,
    breaklines=true,
    numbers=left,
    numberstyle=\tiny\color{gray},
    frame=none,
    xleftmargin=1em
]
# EVOLVE-BLOCK-START
import numpy as np
from scipy.optimize import minimize

def circle_packing21() -> np.ndarray:
    """
    Places 21 non-overlapping circles inside a rectangle of perimeter 4 (w+h=2)
    to maximize the sum of their radii.
    """
    n_circles = 21
    best_sum = 0
    best_circles = np.zeros((n_circles, 3))
    
    # Seeds for reproducibility
    seeds = [42, 123, 999, 7, 88]
    
    def objective(vars):
        # vars: [w, x0, y0, r0, x1, y1, r1, ...]
        # We want to maximize sum(radii), so minimize -sum(radii)
        radii = vars[3::3]
        return -np.sum(radii)

    def constraints(vars, n):
        w = vars[0]
        h = 2.0 - w
        coords = vars[1:]
        res = []
        
        # Boundary constraints: x-r >= 0, x+r <= w, y-r >= 0, y+r <= h
        for i in range(n):
            x, y, r = coords[3*i], coords[3*i+1], coords[3*i+2]
            res.append(x - r)
            res.append(w - (x + r))
            res.append(y - r)
            res.append(h - (y + r))
            res.append(r) # r > 0
            
        # Non-overlap constraints: dist >= r1 + r2
        for i in range(n):
            xi, yi, ri = coords[3*i], coords[3*i+1], coords[3*i+2]
            for j in range(i + 1, n):
                xj, yj, rj = coords[3*j], coords[3*j+1], coords[3*j+2]
                # (xi-xj)**2 + (yi-yj)**2 - (ri+rj)**2 >= 0
                res.append(np.sqrt((xi-xj)**2 + (yi-yj)**2 + 1e-9) - (ri + rj))
        return np.array(res)

    # Optimization Bounds
    # w: [0.5, 1.5], x: [0, 1.5], y: [0, 1.5], r: [0, 0.5]
    bounds = [(0.5, 1.5)] + [(0, 1.5), (0, 1.5), (0, 0.5)] * n_circles

    for seed in seeds:
        np.random.seed(seed)
        
        # Initialization: Start with a rectangle near square and a grid-ish layout
        w_init = 1.0
        h_init = 1.0
        
        # Create a jittered grid for 21 circles (approx 4x5 or 5x5)
        # We'll place them and assign a small initial radius
        x_init = np.linspace(0.1, 0.9, 5)
        y_init = np.linspace(0.1, 0.9, 5)
        grid_x, grid_y = np.meshgrid(x_init, y_init)
        pts = np.vstack([grid_x.ravel(), grid_y.ravel()]).T
        np.random.shuffle(pts)
        pts = pts[:n_circles]
        
        r_init = 0.05 * np.ones((n_circles, 1))
        initial_vars = np.zeros(1 + 3 * n_circles)
        initial_vars[0] = w_init
        for i in range(n_circles):
            initial_vars[1 + 3*i] = pts[i, 0]
            initial_vars[1 + 3*i + 1] = pts[i, 1]
            initial_vars[1 + 3*i + 2] = r_init[i]

        # Optimization
        res = minimize(
            objective,
            initial_vars,
            method='SLSQP',
            constraints={'type': 'ineq', 'fun': lambda v: constraints(v, n_circles)},
            bounds=bounds,
            options={'maxiter': 150, 'ftol': 1e-6}
        )
        
        if res.success or res.fun < 0:
            current_sum = -res.fun
            if current_sum > best_sum:
                best_sum = current_sum
                # Reshape result
                final_w = res.x[0]
                final_coords = res.x[1:].reshape((n_circles, 3))
                best_circles = final_coords

    return best_circles

# EVOLVE-BLOCK-END
\end{lstlisting}

\end{tcolorbox}
\captionof{figure}{Best program discovered by \modelshinka with Gemini-3-Flash in  Circle Packing (Rect).}
\label{fig:case_study_prism}
\end{center}

\begin{center}
\centering
\begin{tcolorbox}[
    colback=cyan!4,
    colframe=cyan!40!black,
    breakable,
    fonttitle=\bfseries\large,
    arc=1mm
]

\begin{lstlisting}[
    language=python,
    basicstyle=\ttfamily\scriptsize,
    keywordstyle=\color{blue}\bfseries,
    commentstyle=\color{green!50!black}\itshape,
    stringstyle=\color{orange!70!black},
    showstringspaces=false,
    breaklines=true,
    numbers=left,
    numberstyle=\tiny\color{gray},
    frame=none,
    xleftmargin=1em
]
import re
import json
from typing import Callable, Tuple
# EVOLVE-BLOCK-START
class Agent:
    def __init__(self, query_llm: Callable, temperature: float = 0.0):
        # Temperature 0.0 is critical for strict instruction following and reproducibility
        self.query_llm = query_llm
        self.temperature = temperature

    def forward(self, prompt: str) -> Tuple[str, float]:
        system_prompt = (
            "You are a precision instruction-following engine. You must satisfy ALL constraints perfectly.\n"
            "Use the following internal reasoning protocol before providing the response:\n\n"
            "1. <PLANNING_SCRATCHPAD>:\n"
            "   - List all explicit constraints (length, format, forbidden words, etc.).\n"
            "   - FORBIDDEN CONTENT: Identify any forbidden words/topics and assign them IDs (e.g., [F1], [F2]). NEVER type the actual forbidden words in this scratchpad.\n"
            "   - STRUCTURE: Define the number of sentences/paragraphs. For each, state the intended first word and last word (Bookending).\n"
            "   - COUNTING: If a word/character count is required, perform a preliminary count here.\n"
            "2. <FINAL_RESPONSE>: Provide the output that satisfies every constraint. Do not include any meta-talk or explanations.\n\n"
            "EXAMPLE:\n"
            "USER: Write a 4-word sentence about a cat without using the word 'meow'.\n"
            "<PLANNING_SCRATCHPAD>:\n"
            "- Topic: Cat\n"
            "- Length: Exactly 4 words\n"
            "- Forbidden: [F1] (the 'm' word)\n"
            "- Plan: Sentence 1 starts with 'The', ends with 'sleeps'.\n"
            "<FINAL_RESPONSE>:\n"
            "The fluffy cat sleeps.\n"
            "END_EXAMPLE"
        )
        
        user_input = (
            f"REQUEST:\n{prompt}\n\n"
            "Strictly follow the <PLANNING_SCRATCHPAD> and <FINAL_RESPONSE> protocol."
        )

        response, cost = self.query_llm(
            prompt=user_input, 
            system=system_prompt, 
            temperature=self.temperature
        )
        
        # Robust extraction focusing on the <FINAL_RESPONSE> tag
        final_output = ""
        if "<FINAL_RESPONSE>" in response:
            # Take the content after the last occurrence of the tag to avoid issues if the model repeats it
            parts = response.split("<FINAL_RESPONSE>")
            final_output = parts[-1].strip()
            # Remove closing tag if present
            if "</FINAL_RESPONSE>" in final_output:
                final_output = final_output.split("</FINAL_RESPONSE>")[0].strip()
        else:
            # Fallback for models that might use lowercase or skip tags
            if "<FINAL_RESPONSE" in response.upper():
                tag_start = response.upper().find("<FINAL_RESPONSE")
                content_start = response.find(">", tag_start) + 1
                final_output = response[content_start:].strip()
            else:
                # If tags are totally missing, take the last block of text
                parts = response.split('\n\n')
                final_output = parts[-1].strip()

        # Final cleanup of common LLM artifacts
        # Remove markdown code blocks
        final_output = re.sub(r'^```\w*\n', '', final_output)
        final_output = re.sub(r'```$', '', final_output).strip()
        
        # Remove any leading colons or labels the model might have prepended
        final_output = re.sub(r'^(Result|Output|Answer|Response):\s*', '', final_output, flags=re.IGNORECASE).strip()
        
        # Remove any trailing "END_EXAMPLE" or similar hallucinated markers
        final_output = re.sub(r'\n?END_EXAMPLE$', '', final_output).strip()
        
        return final_output, cost
# EVOLVE-BLOCK-END
\end{lstlisting}

\end{tcolorbox}
\captionof{figure}{Best program discovered by \modelshinka with Gemini-3-Flash in  IFBench.}
\label{fig:case_study_prism}
\end{center}


\end{document}